\documentclass{article}

\PassOptionsToPackage{authoryear, round}{natbib}

\usepackage[dblblindworkshop, final]{coml_2025}
\workshoptitle{Constrained Optimization for Machine Learning (COML)}

\usepackage[utf8]{inputenc}
\usepackage[T1]{fontenc}
\usepackage{amsmath, amssymb, amsthm}
\usepackage{graphicx}
\usepackage{subfigure}
\usepackage{enumitem}
\usepackage{xcolor}
\usepackage{caption}
\setlist{nosep}
\usepackage{dblfloatfix}
\usepackage{algorithm}
\usepackage[table]{xcolor} 
\usepackage{algpseudocode}
\usepackage{booktabs,makecell,multirow,xcolor}
\usepackage{tabularx}
\usepackage{ragged2e}
\algrenewcommand\algorithmicindent{0.8em}
\algrenewcommand\textproc{}

\theoremstyle{plain}

\theoremstyle{definition}

\theoremstyle{remark}

\usepackage[urlcolor=blue]{hyperref}
\usepackage[capitalize,noabbrev]{cleveref}

\usepackage{sectsty}
\sectionfont{\vspace{-0.6em}} 
\subsectionfont{\vspace{-0.5em}} 
\usepackage{adjustbox}

\title{SIMU: Selective Influence Machine Unlearning}

\author{
Anu Agarwal\thanks{Equal contribution.} \quad
Mihir Pamnani\footnotemark[1] \quad
Dilek Hakkani-Tur \\
University of Illinois Urbana-Champaign \\
\texttt{\{anua2,pamnani3,dilekh\}@illinois.edu}
}

\begin{document}

\maketitle

\begin{abstract}
The undesired memorization of sensitive information by Large Language Models (LLMs) has emphasized the need for safety mechanisms that can regulate model behavior. This has led to the development of machine unlearning techniques that enable models to precisely forget sensitive and unwanted information. For machine unlearning, first-order and second-order optimizer-based methods have shown significant progress in enabling LLMs to forget targeted information. However, in doing so, these approaches often compromise the model’s original capabilities, resulting in unlearned models that struggle to retain their prior knowledge and overall utility \citep{liu2024rethinkingmachineunlearninglarge}. To address this, we propose Selective Influence Machine Unlearning (SIMU), a two-step framework that enhances second-order optimizer-based unlearning by selectively updating only the \textit{critical neurons} responsible for encoding the forget-set. By constraining updates to these targeted neurons, SIMU achieves comparable unlearning efficacy while substantially outperforming current methods in retaining the model’s original knowledge.
\end{abstract}

\section{Introduction}
Autoregressive Large Language Models (LLMs) have made tremendous progress on natural language tasks since the introduction of the transformer architecture \citep{vaswani2023attentionneed}. However, their ability to memorize large portions of training data has raised significant concerns regarding data privacy, intellectual property rights, and the influence of undesired content. With increasing emphasis on data protection rights and AI safety practices, machine unlearning has emerged as a salient research direction. Techniques in this domain enable LLMs to unlearn targeted content while preserving overall effectiveness. We formalize machine unlearning as a constrained optimization problem during training: given a target model with a retain-set $\mathcal{D}_r$ and a forget-set $\mathcal{D}_f$, the objective is to construct an unlearned model that preserves knowledge from $\mathcal{D}_r$ while eliminating the influence of $\mathcal{D}_f$.

Given the high cost of retraining models from scratch, fine-tuning under a predefined unlearning objective has become the primary strategy for LLM unlearning. For instance, classical gradient ascent-based fine-tuning methods are prone to over-forgetting, which can degrade model utility \citep{zhang2024negativepreferenceoptimizationcatastrophic}. In contrast, less aggressive methods, such as fine-tuning only on the retain-set, may result in under-forgetting, thereby failing to fully erase the influence of the forget-set data \citep{yao2024largelanguagemodelunlearning}. A widely adopted solution is a regularized optimization objective that balances unlearning efficacy and utility preservation. This principle underlies methods such as Gradient Difference (GradDiff) \citep{pmlr-v199-liu22a}, Preference Optimization (PO) \citep{eldan2023whosharrypotterapproximate, maini2024tofutaskfictitiousunlearning}, and Negative Preference Optimization (NPO) \citep{zhang2024negativepreferenceoptimizationcatastrophic}. In localization-informed unlearning, techniques leverage model internals to apply fine-tuning selectively to a subset of components (e.g., layers or neurons) most relevant to the unlearning objective \citep{yu-etal-2023-unlearning, wu2023depndetectingeditingprivacy}. On the other hand, influence-function-based methods \citep{koh2020understandingblackboxpredictionsinfluence} attempt to update model parameters in a single shot for effective unlearning. SOUL \citep{jia2024soulunlockingpowersecondorder} re-frames influence-based unlearning as an iterative optimization process, where updates informed by second-order derivatives yield strong unlearning performance while preserving much of the model's original capabilities. To our knowledge, no prior work has explicitly addressed the issue of minimizing Hessian approximation errors in second-order influence-based unlearning with localization-informed techniques. To bridge this gap, we propose \textbf{Selective Influence Machine Unlearning (SIMU)}, a novel framework designed to enhance second-order (SO) optimization for unlearning. Our technical contributions are as follows:
\begin{itemize}[topsep=2pt, partopsep=0pt, itemsep=2pt, parsep=0pt]
 \item We investigate the role of neurons as the primary model unit for updates in LLM unlearning, explicitly in aggressive unlearning regimes such as Gradient Difference.
 \item We propose \textbf{SIMU}, a two-step, second-order unlearning framework that improves unlearning through an intelligent masking strategy applied during fine-tuning.
\end{itemize}

\section{Methodology}

\captionsetup[figure]{font=normal}
\begin{figure}
    \centering
    \includegraphics[width=0.8\linewidth]{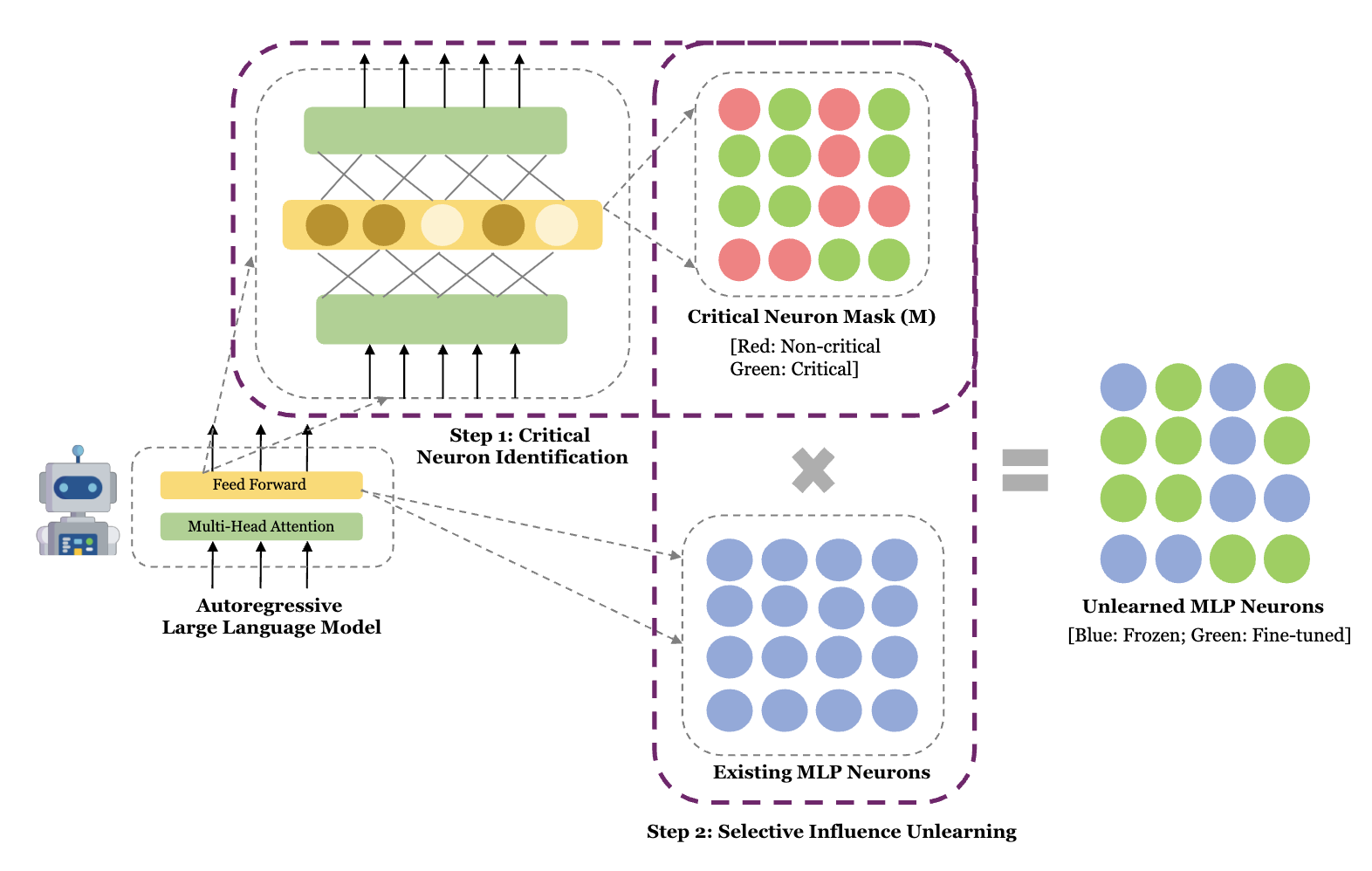}
    \caption{Overview of the SIMU framework. First, we build a Critical Neuron Mask by identifying MLP neurons associated with forget-set knowledge, and then perform selective unlearning on these critical neurons and the attention layers, while keeping the remaining parameters frozen.}
    \label{fig:pipeline}
\end{figure}

\subsection{Critical Neuron Identification}
\label{subsection:DEPN}
The first step in our proposed framework (Figure \ref{fig:pipeline}) involves identifying the \textit{critical neurons} in MLP layers that contribute more to encode the information to be forgotten. Numerous attempts have been made to localize parts of a language model that store information relevant to specific data samples \citep{meng2023locatingeditingfactualassociations}. To ensure precise control while editing the information encoded in the model, neuron-level granularity is considered to be suitable for machine unlearning. As underlined in \citep{meng2023locatingeditingfactualassociations, meng2022mass}, the MLP in every layer of the transformer acts as key-value memory and stores the factual knowledge possessed by these models.  While attention layers capture long-range contextual relationships between tokens, MLP
layers are responsible for feature transformation. Thus, extending the Privacy Neuron Detector \citep{wu2023depndetectingeditingprivacy} for masked language models, we propose a gradient-aggregation approach to calculate the forget-set attribution score for MLP neurons in autoregressive language models.

 Given a forget-set \(D\) of question–answer pairs, we convert each pair to multiple next-token prediction-style samples.
 For each neuron \(w_l^k\) ($k$-th neuron of the $l$-th MLP down-sample layer) and each sample \(i\), let \(\beta_{l,i}^k\) denote the neuron’s original activation on that sample.  We measure a neuron’s contribution for a sample by controlled scaling of its activation from \(0\) to \(\beta_{l,i}^k\) in \(m\) evenly spaced steps (Eq. \ref{eq:activation_steps}) and computing the per-step loss (Eq. \ref{eq:loss})
 
\begin{subequations}
\noindent
\begin{minipage}{0.48\textwidth}
\begin{equation}
a_{j,l,i}^k = \frac{j}{m}\,\beta_{l,i}^k, \quad j=1,\dots,m
\label{eq:activation_steps}
\end{equation}
\end{minipage}\hfill
\begin{minipage}{0.48\textwidth}
\begin{equation}
L_i(a_{j,l,i}^k) = -\log P(y_i \mid X_i, a_{j,l,i}^k)
\label{eq:loss}
\end{equation}
\end{minipage}
\end{subequations}

where \(P(\cdot)\) denotes the model’s autoregressive token probability when the neuron’s activation is set to \(a_{j,l,i}^k\). We then take the gradient of this loss with respect to the injected activation and aggregate these gradients across \(m\) steps and all samples to obtain the neuron's attribution score:
\begin{equation}
\operatorname{Att}(w_l^k)
\;=\;
\frac{1}{m}\sum_{j=1}^m\sum_{i=1}^{|D|}
\beta_{l,i}^k \,\frac{\partial L_i\!\big(a_{j,l,i}^k\big)}{\partial a_{j,l,i}^k}.
\label{eq:attribution}
\end{equation}
Finally, we convert these attribution scores to a per-layer binary mask by thresholding.  Let \(M_l=\max_k \operatorname{Att}(w_l^k)\). A neuron \(w_l^k\) is declared \emph{critical} iff
$\operatorname{Att}(w_l^k) > t \cdot M_l$,
where \(t\in(0,1]\) is a tunable parameter that controls the fraction of neurons selected per layer.  The resulting binary mask introduces structural sparsity \citep{zheng2024learnefficientbuildstructured} in MLP updates and is used to guide the selective unlearning in the next phase.

\subsection{Selective Influence Unlearning}
\label{subsection:SOUL}

In the second phase of SIMU, we perform targeted unlearning to satisfy the constrained optimization problem by fine-tuning with the Sophia optimizer \citep{liu2024sophiascalablestochasticsecondorder} inside a second-order iterative framework \citep{jia2024soulunlockingpowersecondorder}. To localize update effects, we freeze all parameters besides the attention projection layers and MLP down-sample layer. Within MLP, we restrict parameter changes to the critical neurons identified by the binary mask \(\mathbf{M}\) from the first phase (see Figure~\ref{fig:pipeline}). Intuitively, preserving attention updates and having sparse MLP updates, preserves the model's sequence modeling capabilities while allowing precise correction where the forget-set signal is concentrated.

Each Newton-like Sophia step updates parameters as a scaled, clipped quasi-Newton step:
\begin{equation}
\theta_{t+1} \;=\; \theta_t \;-\; \eta_t \cdot \operatorname{clip}\!\Big(\frac{m_t}{\max\{\gamma H_t,\epsilon\}},\,1\Big),
\label{eq:sophia_update}
\end{equation}
where \(\eta_t>0\) is the learning rate, \(m_t\) is the EMA of the first moments, \(H_t\) is an EMA approximation of the diagonal (Gauss–Newton), \(\gamma>0\) is a damping factor for numerical stability, and \(\epsilon>0\) prevents division by zero. The first and second-moment EMAs are computed as
\begin{equation}
m_t \;=\; \beta_1 m_{t-1} + (1-\beta_1) g_t,
\qquad
H_t \;=\; \beta_2 H_{t-1} + (1-\beta_2) g_t^2,
\label{eq:fo_ho}
\end{equation}
where \(g_t=\nabla_{\theta} \mathcal{L}_t(\theta_t)\) is the current gradient, and \(\beta_1,\beta_2\in(0,1)\) are momentum coefficients.

To ensure updates only affect critical neurons, we apply the binary layer-wise mask \(\mathbf{M}\) (with complement \(\bar{\mathbf{M}}=1-\mathbf{M}\)) at three precise points in each iteration: (i) after computing the first-moment EMA, (ii) after computing the curvature EMA, and (iii) after forming the full parameter update. Concretely,
\begin{equation}
\begin{aligned}
m'_t &\;=\; \beta_1 m_{t-1} + (1-\beta_1) g_t,\\
m_t &\;=\; \mathbf{M}\odot m'_t \;+\; \bar{\mathbf{M}}\odot m_{t-1},
\end{aligned}
\label{eq:masked_mt}
\end{equation}
\begin{equation}
\begin{aligned}
H'_t &\;=\; \beta_2 H_{t-1} + (1-\beta_2) g_t^2,\\
H_t &\;=\; \mathbf{M}\odot H'_t \;+\; \bar{\mathbf{M}}\odot H_{t-1},
\end{aligned}
\label{eq:masked_Ht}
\end{equation}
and
\begin{equation}
\theta_{t} \;=\; \mathbf{M}\odot \theta'_{t} \;+\; \bar{\mathbf{M}}\odot \theta_{t-1},
\label{eq:masked_theta}
\end{equation}
where \(\theta'_{t}\) denotes the complete Sophia update after including the weight decay and applying \eqref{eq:sophia_update} elementwise. Here, \(\odot\) denotes element-wise multiplication, and all masking is restricted to parameters inside MLP modules; parameters of other modules remain equal to their previous values.

This masked second-order fine-tuning confines the parameter updates to neurons that our attribution procedure identified as \textit{critical}, thereby producing a controlled unlearning step that (i) removes the targeted information effectively and (ii) minimizes collateral damage to retained knowledge. As discussed in Section~\ref{subsection:influence-unlearning}, the Sophia-based masked update closely resembles an influence-function-style correction: it applies a (clipped) approximate Newton step concentrated on the neurons most responsible for the forget-set loss, yielding practical and stable unlearning in large autoregressive models.

{\setlength{\parskip}{0pt}

\section{Experiments}
\paragraph{Experimentation Setup:}
We evaluate SIMU on two unlearning benchmarks. \textbf{(1) TOFU:} a fictitious-unlearning task \citep{maini2024tofutaskfictitiousunlearning} which consists of synthetic author profiles. \textbf{(2) LUME:} a multi-faceted benchmark from \citep{ramakrishna2025lumellmunlearningmultitask} which consists of three subtasks: (a) long-form creative texts, (b) short-form fake PII (e.g., contact details, SSNs), and (c) real documents drawn from training data. For both benchmarks, we run experiments on LLaMA2-7B \citep{touvron2023llama2openfoundation} and OLMo-1B \citep{groeneveld2024olmoacceleratingsciencelanguage}, to measure unlearning effectiveness and utility preservation. As baselines, we compare against Gradient-Difference implemented with both first-order (FO-GradDiff \citep{pmlr-v199-liu22a}) and second-order (SO-GradDiff \citep{jia2024soulunlockingpowersecondorder}) optimizers. Hyperparameter configurations used for all experiments can be found in Appendix \S\ref{section:hyperparameters}.

\paragraph{Results:} As shown in Tables \ref{tab:tofu_main_results} and \ref{tab:lume_main_results}, our method consistently outperforms prior baselines on both TOFU and LUME, with substantially higher model utility while maintaining comparable unlearning efficacy. For OLMo-1B, we observe a 1–2\% improvement in utility relative to SO-GradDiff, and for LLaMA2-7B the improvement is about 5–6\%. Since, SIMU-GradDiff can be viewed as an improved SO-GradDiff that restricts updates to a subset of MLP neurons, these results indicate that targeting specific model components better preserves utility. At the same time, we observe that the magnitude of improvement varies between model architectures and sizes: gains are larger for the LLaMA2-7B model than for OLMo-1B. We attribute this to the fact that, across tasks, LLaMa2-7B has a more concentrated forget-set signal in a small set of critical neurons compared to OLMo-1B (see discussion in Section \ref{section:hyperparameters}). Overall, the improved aggregate scores for SIMU-GradDiff support our main claim: for aggressive unlearning approaches such as GradDiff, combining full attention updates with structured, sparse MLP updates provides the ideal balance between effective forgetting and preserving overall model utility.

\captionsetup{font=scriptsize}
\begin{table*}[t]
  \caption{Overview of unlearning performance on TOFU. The aggregate score combines forgetting and retention performance, computed as the mean of ($1 -$ unlearning efficacy metrics), utility metrics on retain-set, and ($1 -$ MIA).}
  \label{tab:tofu_main_results}
  \centering
  \resizebox{\textwidth}{!}{%
    \begin{tabular}{@{}llcccccccc@{}}
      \toprule
      \textbf{Model} & \textbf{Approach} & \makecell{\textbf{Aggregate } \\
      \textbf{Score} ($\uparrow$)}
      & \multicolumn{3}{c}{\textbf{Unlearning Efficacy}}
      & \multicolumn{4}{c}{\textbf{Utility}} \\
      \cmidrule(lr){4-6} \cmidrule(lr){7-10}
      & & 
      & \makecell{\textbf{Exact Match} \\ \textbf{- Forget} \(\downarrow\)}
      & \makecell{\textbf{Rouge-L} \\ \textbf{- Forget} \(\downarrow\)}
      & \makecell{\textbf{MIA}  \(\downarrow\)}
      & \makecell{\textbf{Exact Match} \\ \textbf{- Retain} \(\uparrow\)}
      & \makecell{\textbf{Rouge-L} \\ \textbf{- Retain} \(\uparrow\)}
      & \makecell{\textbf{Exact Match} \\ \textbf{- World Facts} \(\uparrow\)}
      & \makecell{\textbf{Rouge-L} \\ \textbf{- World Facts} \(\uparrow\)} \\
      \midrule

      \multirow{4}{*}{\makecell{LLaMA2-7B}} & Original
        & 0.4437
        & 85.25\% & 0.9796 & 0.7894
        & 85.75\% & 0.9825 & 86.32\% & 0.8960 \\
        \cmidrule(lr){2-10}

      & FO-GradDiff
        & 0.4738
        & 72.75\% & 0.5174 & 0.7627
        & 76.50\% & 0.6115 & 79.49\% & 0.8462 \\

      & SO-GradDiff
        & 0.7957
        & \textbf{10.25\%} & \textbf{0.0221} & \textbf{0.2156}
        & 72.25\% & 0.5960 & 82.05\% & 0.8675 \\

      & SIMU-GradDiff (Ours)
        & \textbf{0.7963}
        & 20\% & 0.0241 & 0.2440
        & \textbf{78.00\%} & \textbf{0.6694} & \textbf{82.90\%} & \textbf{0.8703} \\

    \cmidrule(lr){1-10}

    \multirow{4}{*}{OLMo-1B} & Original
      & 0.4227
      & 77.50\% & 0.8503 & 0.7727
      & 78.75\% & 0.8239 & 48.71\% & 0.5537 \\
      \cmidrule(lr){2-10}
    
    & FO-GradDiff
      & 0.7059
      & 26.50\% & 0.0214 & 0.1957
      & 63.00\% & 0.3814 & 0.85\% & 0.0185 \\
    
    & SO-GradDiff
      & 0.8235
      & 22.75\% & 0.0077 & \textbf{0.1889}
      & \textbf{78.00\%} & 0.7614 & 38.46\% & 0.4518 \\
    
    & SIMU-GradDiff (Ours)
      & \textbf{0.8438}
      & \textbf{10.25\%} & \textbf{0.0029} & 0.1923
      & 75.50\% & \textbf{0.7616} & \textbf{42.74\%} & \textbf{0.4896} \\

    \bottomrule
    \end{tabular}%
    }
\end{table*}

\begin{table*}[t]
  \caption{\fontsize{7}{8.6}\selectfont Overview of unlearning performance across three tasks on LUME.  Each cell except MIA represents
(Regurgitation Score; Knowledge Score) where Regurgitation Score and Knowledge Score are similar to ROUGE-L and Exact-Match
respectively. The aggregate score combines forgetting and retention performance, computed as the mean of ($1 -$ unlearning efficacy metrics), utility metrics on retain-set, and ($1 -$ MIA).}
  \label{tab:lume_main_results}
  \centering
  \begin{adjustbox}{max width=\textwidth,center}
    \begin{tabular}{@{}llccccccccccc@{}}
      \toprule
      \textbf{Model} & \textbf{Approach} & \makecell{\textbf{Aggregate} \\  \textbf{Score} ($\uparrow$)}
      & \multicolumn{5}{c}{\textbf{Unlearning Efficacy}}
      & \multicolumn{4}{c}{\textbf{Utility}} \\
      \cmidrule(lr){4-8} \cmidrule(lr){9-12}

      & & 
      & \makecell{\textbf{Overall} \\ \textbf{(Reg \(\downarrow\); Kno \(\downarrow\))}}
      & \makecell{\textbf{Task-1} \\ \textbf{(Reg \(\downarrow\); Kno \(\downarrow\))}}
      & \makecell{\textbf{Task-2} \\ \textbf{(Reg \(\downarrow\); Kno \(\downarrow\))}}
      & \makecell{\textbf{Task-3} \\ \textbf{(Reg \(\downarrow\); Kno \(\downarrow\))}}
      & \makecell{\textbf{MIA} \\ \(\downarrow\)}
      & \makecell{\textbf{Overall} \\ \textbf{(Reg \(\uparrow\); Kno \(\uparrow\))}}
      & \makecell{\textbf{Task-1} \\ \textbf{(Reg \(\uparrow\); Kno \(\uparrow\))}}
      & \makecell{\textbf{Task-2} \\ \textbf{(Reg \(\uparrow\); Kno \(\uparrow\))}}
      & \makecell{\textbf{Task-3} \\ \textbf{(Reg \(\uparrow\); Kno \(\uparrow\))}} \\

    \midrule
      \multirow{4}{*}{\makecell{LLaMA2-7B}} & Original & 0.504
        & 0.9885 / 0.7984 & 1.00 / 0.9137 & 0.9575 / 0.7333 & 1.00 / 0.9231 & 0.458
        & 0.9859 / 0.7741 & 1.00 / 0.8241 & 0.9600 / 0.7280 & 0.9957 / 0.9111 \\
        \cmidrule(lr){2-12}

      & FO-GradDiff & 0.586
        & 0.5566 / 0.2029 & 0.7046 / 0.1121 & 0.6071 / 0.2000 & 0.4027 / 0.2867 & 0.4769
        & 0.7825 / 0.4463 & \textbf{0.9134} / \textbf{0.6483} & 0.7032 / 0.3220 & 0.7524 / 0.7703 \\

      & SO-GradDiff & 0.607
        & 0.0187 / 0.00 & 0.00 / 0.00 & 0.0674 / 0.00 & 0.0012 / 0.00 & 0.5391
        & 0.7714 / 0.6212 & 0.8176 / 0.5495 & 0.7102 / 0.5780 & 0.7857 / 0.8296 \\

      & SIMU-GradDiff (Ours) & \textbf{0.659}
        & \textbf{0.0025} / \textbf{0.00} & \textbf{0.00} / \textbf{0.00}
        & \textbf{0.0100} / \textbf{0.00} & \textbf{0.00} / \textbf{0.00} & \textbf{0.5333}
        & \textbf{0.8295} / \textbf{0.7149} & 0.8180 / 0.4725 & \textbf{0.7885} / \textbf{0.7260} & \textbf{0.8672} / \textbf{0.8370} \\

      \midrule
      \multirow{4}{*}{\makecell{OLMo-1B}} & Original & 0.211
        & 0.9837 / 0.8687 & 1.00 / 1.00 & 0.9399 / 0.80 & 1.00 / 0.9930 & 1.0
        & 0.9857 / 0.8623 & 1.00 / 1.00 & 0.9534 / 0.80 & 1.00 / 1.00 \\
        \cmidrule(lr){2-12}

      & FO-GradDiff & 0.512
        & 0.1605 / 0.3681 & 0.0564 / 0.25 & 0.4616 / 0.4646 & 0.0430 / 0.1399 & 0.9776
        & 0.3854 / 0.7314 & 0.1783 / 0.7582 & 0.7274 / 0.7420 & 0.2716 / 0.6741 \\

      & SO-GradDiff & 0.728
        & 0.0055 / 0.0 & 0.0029 / 0.0
        & \textbf{0.0033} / \textbf{0.0} & 0.0091 / 0.0 & 0.7035
        & 0.9244 / 0.8499 & \textbf{0.9781} / \textbf{0.9670} & 0.8271 / 0.7900 & 0.9602 / 0.9926 \\

      & SIMU-GradDiff (Ours) & \textbf{0.740}
        & \textbf{0.0015} / \textbf{0.0} & \textbf{0.0} / \textbf{0.0}
        & 0.0050 / 0.0 & \textbf{0.0004} / \textbf{0.0} & \textbf{0.6889}
        & \textbf{0.9365} / \textbf{0.8540} & 0.9779 / 0.9670 & \textbf{0.8549} / \textbf{0.7960} & \textbf{0.9690} / \textbf{0.9926} \\

      \bottomrule
    \end{tabular}%
  \end{adjustbox}
\end{table*}

\section{Conclusion}
In this paper, we introduce SIMU, a novel selective influence–based machine unlearning framework designed specifically for autoregressive language models. Our approach comprises two key components: first, a sophisticated neuron identification mechanism that pinpoints critical neurons with substantial contributions to forget-set information; and second, a targeted second-order unlearning procedure that operates exclusively on these identified neurons. Through extensive empirical evaluations, we demonstrate SIMU’s effectiveness in selectively eliminating forget-set information while significantly improving the model’s performance on retain-set tasks. The experimental results strongly validate our hypothesis that controlled, targeted unlearning updates can minimize the impact of approximation errors in second-order influence unlearning techniques and successfully balance the dual objectives of information removal and utility preservation.

\bibliographystyle{plainnat}   
\bibliography{custom}

\begin{thebibliography}{24}
\providecommand{\natexlab}[1]{#1}
\providecommand{\url}[1]{\texttt{#1}}
\expandafter\ifx\csname urlstyle\endcsname\relax
  \providecommand{\doi}[1]{doi: #1}\else
  \providecommand{\doi}{doi: \begingroup \urlstyle{rm}\Url}\fi

\bibitem[Eldan and Russinovich(2023)]{eldan2023whosharrypotterapproximate}
Ronen Eldan and Mark Russinovich.
\newblock Who's harry potter? approximate unlearning in llms, 2023.
\newblock URL \url{https://arxiv.org/abs/2310.02238}.

\bibitem[Fan et~al.(2024)Fan, Liu, Zhang, Wong, Wei, and Liu]{fan2024salunempoweringmachineunlearning}
Chongyu Fan, Jiancheng Liu, Yihua Zhang, Eric Wong, Dennis Wei, and Sijia Liu.
\newblock Salun: Empowering machine unlearning via gradient-based weight saliency in both image classification and generation, 2024.
\newblock URL \url{https://arxiv.org/abs/2310.12508}.

\bibitem[Groeneveld et~al.(2024)Groeneveld, Beltagy, Walsh, Bhagia, Kinney, Tafjord, Jha, Ivison, Magnusson, Wang, Arora, Atkinson, Authur, Chandu, Cohan, Dumas, Elazar, Gu, Hessel, Khot, Merrill, Morrison, Muennighoff, Naik, Nam, Peters, Pyatkin, Ravichander, Schwenk, Shah, Smith, Strubell, Subramani, Wortsman, Dasigi, Lambert, Richardson, Zettlemoyer, Dodge, Lo, Soldaini, Smith, and Hajishirzi]{groeneveld2024olmoacceleratingsciencelanguage}
Dirk Groeneveld, Iz~Beltagy, Pete Walsh, Akshita Bhagia, Rodney Kinney, Oyvind Tafjord, Ananya~Harsh Jha, Hamish Ivison, Ian Magnusson, Yizhong Wang, Shane Arora, David Atkinson, Russell Authur, Khyathi~Raghavi Chandu, Arman Cohan, Jennifer Dumas, Yanai Elazar, Yuling Gu, Jack Hessel, Tushar Khot, William Merrill, Jacob Morrison, Niklas Muennighoff, Aakanksha Naik, Crystal Nam, Matthew~E. Peters, Valentina Pyatkin, Abhilasha Ravichander, Dustin Schwenk, Saurabh Shah, Will Smith, Emma Strubell, Nishant Subramani, Mitchell Wortsman, Pradeep Dasigi, Nathan Lambert, Kyle Richardson, Luke Zettlemoyer, Jesse Dodge, Kyle Lo, Luca Soldaini, Noah~A. Smith, and Hannaneh Hajishirzi.
\newblock Olmo: Accelerating the science of language models, 2024.
\newblock URL \url{https://arxiv.org/abs/2402.00838}.

\bibitem[Hampel(1974)]{hampel1974influence}
F.~R. Hampel.
\newblock The influence curve and its role in robust estimation.
\newblock \emph{Journal of the American Statistical Association}, 69\penalty0 (346):\penalty0 383--393, 1974.
\newblock \doi{10.1080/01621459.1974.10482962}.

\bibitem[Jia et~al.(2024)Jia, Zhang, Zhang, Liu, Runwal, Diffenderfer, Kailkhura, and Liu]{jia2024soulunlockingpowersecondorder}
Jinghan Jia, Yihua Zhang, Yimeng Zhang, Jiancheng Liu, Bharat Runwal, James Diffenderfer, Bhavya Kailkhura, and Sijia Liu.
\newblock Soul: Unlocking the power of second-order optimization for llm unlearning, 2024.
\newblock URL \url{https://arxiv.org/abs/2404.18239}.

\bibitem[Koh and Liang(2020)]{koh2020understandingblackboxpredictionsinfluence}
Pang~Wei Koh and Percy Liang.
\newblock Understanding black-box predictions via influence functions, 2020.
\newblock URL \url{https://arxiv.org/abs/1703.04730}.

\bibitem[Liu et~al.(2022{\natexlab{a}})Liu, Liu, and Stone]{liu2022continuallearningprivateunlearning}
Bo~Liu, Qiang Liu, and Peter Stone.
\newblock Continual learning and private unlearning, 2022{\natexlab{a}}.
\newblock URL \url{https://arxiv.org/abs/2203.12817}.

\bibitem[Liu et~al.(2022{\natexlab{b}})Liu, Liu, and Stone]{pmlr-v199-liu22a}
Bo~Liu, Qiang Liu, and Peter Stone.
\newblock Continual learning and private unlearning.
\newblock In Sarath Chandar, Razvan Pascanu, and Doina Precup, editors, \emph{Proceedings of The 1st Conference on Lifelong Learning Agents}, volume 199 of \emph{Proceedings of Machine Learning Research}, pages 243--254. PMLR, 22--24 Aug 2022{\natexlab{b}}.
\newblock URL \url{https://proceedings.mlr.press/v199/liu22a.html}.

\bibitem[Liu et~al.(2024{\natexlab{a}})Liu, Li, Hall, Liang, and Ma]{liu2024sophiascalablestochasticsecondorder}
Hong Liu, Zhiyuan Li, David Hall, Percy Liang, and Tengyu Ma.
\newblock Sophia: A scalable stochastic second-order optimizer for language model pre-training, 2024{\natexlab{a}}.
\newblock URL \url{https://arxiv.org/abs/2305.14342}.

\bibitem[Liu et~al.(2024{\natexlab{b}})Liu, Yao, Jia, Casper, Baracaldo, Hase, Yao, Liu, Xu, Li, Varshney, Bansal, Koyejo, and Liu]{liu2024rethinkingmachineunlearninglarge}
Sijia Liu, Yuanshun Yao, Jinghan Jia, Stephen Casper, Nathalie Baracaldo, Peter Hase, Yuguang Yao, Chris~Yuhao Liu, Xiaojun Xu, Hang Li, Kush~R. Varshney, Mohit Bansal, Sanmi Koyejo, and Yang Liu.
\newblock Rethinking machine unlearning for large language models, 2024{\natexlab{b}}.
\newblock URL \url{https://arxiv.org/abs/2402.08787}.

\bibitem[Maini et~al.(2024)Maini, Feng, Schwarzschild, Lipton, and Kolter]{maini2024tofutaskfictitiousunlearning}
Pratyush Maini, Zhili Feng, Avi Schwarzschild, Zachary~C. Lipton, and J.~Zico Kolter.
\newblock Tofu: A task of fictitious unlearning for llms, 2024.
\newblock URL \url{https://arxiv.org/abs/2401.06121}.

\bibitem[Meng et~al.(2022)Meng, Sharma, Andonian, Belinkov, and Bau]{meng2022mass}
Kevin Meng, Arnab~Sen Sharma, Alex Andonian, Yonatan Belinkov, and David Bau.
\newblock Mass-editing memory in a transformer.
\newblock \emph{arXiv preprint arXiv:2210.07229}, 2022.

\bibitem[Meng et~al.(2023)Meng, Bau, Andonian, and Belinkov]{meng2023locatingeditingfactualassociations}
Kevin Meng, David Bau, Alex Andonian, and Yonatan Belinkov.
\newblock Locating and editing factual associations in gpt, 2023.
\newblock URL \url{https://arxiv.org/abs/2202.05262}.

\bibitem[Rafailov et~al.(2024)Rafailov, Sharma, Mitchell, Ermon, Manning, and Finn]{rafailov2024directpreferenceoptimizationlanguage}
Rafael Rafailov, Archit Sharma, Eric Mitchell, Stefano Ermon, Christopher~D. Manning, and Chelsea Finn.
\newblock Direct preference optimization: Your language model is secretly a reward model, 2024.
\newblock URL \url{https://arxiv.org/abs/2305.18290}.

\bibitem[Ramakrishna et~al.(2025)Ramakrishna, Wan, Jin, Chang, Bu, Vinzamuri, Cevher, Hong, and Gupta]{ramakrishna2025lumellmunlearningmultitask}
Anil Ramakrishna, Yixin Wan, Xiaomeng Jin, Kai-Wei Chang, Zhiqi Bu, Bhanukiran Vinzamuri, Volkan Cevher, Mingyi Hong, and Rahul Gupta.
\newblock Lume: Llm unlearning with multitask evaluations, 2025.
\newblock URL \url{https://arxiv.org/abs/2502.15097}.

\bibitem[Schulman et~al.(2017{\natexlab{a}})Schulman, Levine, Moritz, Jordan, and Abbeel]{schulman2017trustregionpolicyoptimization}
John Schulman, Sergey Levine, Philipp Moritz, Michael~I. Jordan, and Pieter Abbeel.
\newblock Trust region policy optimization, 2017{\natexlab{a}}.
\newblock URL \url{https://arxiv.org/abs/1502.05477}.

\bibitem[Schulman et~al.(2017{\natexlab{b}})Schulman, Wolski, Dhariwal, Radford, and Klimov]{schulman2017proximalpolicyoptimizationalgorithms}
John Schulman, Filip Wolski, Prafulla Dhariwal, Alec Radford, and Oleg Klimov.
\newblock Proximal policy optimization algorithms, 2017{\natexlab{b}}.
\newblock URL \url{https://arxiv.org/abs/1707.06347}.

\bibitem[Touvron et~al.(2023)Touvron, Martin, Stone, Albert, Almahairi, Babaei, Bashlykov, Batra, Bhargava, Bhosale, Bikel, Blecher, Ferrer, Chen, Cucurull, Esiobu, Fernandes, Fu, Fu, Fuller, Gao, Goswami, Goyal, Hartshorn, Hosseini, Hou, Inan, Kardas, Kerkez, Khabsa, Kloumann, Korenev, Koura, Lachaux, Lavril, Lee, Liskovich, Lu, Mao, Martinet, Mihaylov, Mishra, Molybog, Nie, Poulton, Reizenstein, Rungta, Saladi, Schelten, Silva, Smith, Subramanian, Tan, Tang, Taylor, Williams, Kuan, Xu, Yan, Zarov, Zhang, Fan, Kambadur, Narang, Rodriguez, Stojnic, Edunov, and Scialom]{touvron2023llama2openfoundation}
Hugo Touvron, Louis Martin, Kevin Stone, Peter Albert, Amjad Almahairi, Yasmine Babaei, Nikolay Bashlykov, Soumya Batra, Prajjwal Bhargava, Shruti Bhosale, Dan Bikel, Lukas Blecher, Cristian~Canton Ferrer, Moya Chen, Guillem Cucurull, David Esiobu, Jude Fernandes, Jeremy Fu, Wenyin Fu, Brian Fuller, Cynthia Gao, Vedanuj Goswami, Naman Goyal, Anthony Hartshorn, Saghar Hosseini, Rui Hou, Hakan Inan, Marcin Kardas, Viktor Kerkez, Madian Khabsa, Isabel Kloumann, Artem Korenev, Punit~Singh Koura, Marie-Anne Lachaux, Thibaut Lavril, Jenya Lee, Diana Liskovich, Yinghai Lu, Yuning Mao, Xavier Martinet, Todor Mihaylov, Pushkar Mishra, Igor Molybog, Yixin Nie, Andrew Poulton, Jeremy Reizenstein, Rashi Rungta, Kalyan Saladi, Alan Schelten, Ruan Silva, Eric~Michael Smith, Ranjan Subramanian, Xiaoqing~Ellen Tan, Binh Tang, Ross Taylor, Adina Williams, Jian~Xiang Kuan, Puxin Xu, Zheng Yan, Iliyan Zarov, Yuchen Zhang, Angela Fan, Melanie Kambadur, Sharan Narang, Aurelien Rodriguez, Robert Stojnic, Sergey Edunov, and Thomas
  Scialom.
\newblock Llama 2: Open foundation and fine-tuned chat models, 2023.
\newblock URL \url{https://arxiv.org/abs/2307.09288}.

\bibitem[Vaswani et~al.(2023)Vaswani, Shazeer, Parmar, Uszkoreit, Jones, Gomez, Kaiser, and Polosukhin]{vaswani2023attentionneed}
Ashish Vaswani, Noam Shazeer, Niki Parmar, Jakob Uszkoreit, Llion Jones, Aidan~N. Gomez, Lukasz Kaiser, and Illia Polosukhin.
\newblock Attention is all you need, 2023.
\newblock URL \url{https://arxiv.org/abs/1706.03762}.

\bibitem[Wu et~al.(2023)Wu, Li, Xu, Dong, Wu, Bian, and Xiong]{wu2023depndetectingeditingprivacy}
Xinwei Wu, Junzhuo Li, Minghui Xu, Weilong Dong, Shuangzhi Wu, Chao Bian, and Deyi Xiong.
\newblock Depn: Detecting and editing privacy neurons in pretrained language models, 2023.
\newblock URL \url{https://arxiv.org/abs/2310.20138}.

\bibitem[Yao et~al.(2024)Yao, Xu, and Liu]{yao2024largelanguagemodelunlearning}
Yuanshun Yao, Xiaojun Xu, and Yang Liu.
\newblock Large language model unlearning, 2024.
\newblock URL \url{https://arxiv.org/abs/2310.10683}.

\bibitem[Yu et~al.(2023)Yu, Jeoung, Kasi, Yu, and Ji]{yu-etal-2023-unlearning}
Charles Yu, Sullam Jeoung, Anish Kasi, Pengfei Yu, and Heng Ji.
\newblock Unlearning bias in language models by partitioning gradients.
\newblock In Anna Rogers, Jordan Boyd-Graber, and Naoaki Okazaki, editors, \emph{Findings of the Association for Computational Linguistics: ACL 2023}, pages 6032--6048, Toronto, Canada, July 2023. Association for Computational Linguistics.
\newblock \doi{10.18653/v1/2023.findings-acl.375}.
\newblock URL \url{https://aclanthology.org/2023.findings-acl.375}.

\bibitem[Zhang et~al.(2024)Zhang, Lin, Bai, and Mei]{zhang2024negativepreferenceoptimizationcatastrophic}
Ruiqi Zhang, Licong Lin, Yu~Bai, and Song Mei.
\newblock Negative preference optimization: From catastrophic collapse to effective unlearning, 2024.
\newblock URL \url{https://arxiv.org/abs/2404.05868}.

\bibitem[Zheng et~al.(2024)Zheng, Bai, Liu, Mao, Chen, Lai, and Prakash]{zheng2024learnefficientbuildstructured}
Haizhong Zheng, Xiaoyan Bai, Xueshen Liu, Z.~Morley Mao, Beidi Chen, Fan Lai, and Atul Prakash.
\newblock Learn to be efficient: Build structured sparsity in large language models, 2024.
\newblock URL \url{https://arxiv.org/abs/2402.06126}.

\end{thebibliography}

\appendix
{\setlength{\parskip}{1pt}
\section{Related Work}
In this section, we briefly discuss four categories of established techniques for machine unlearning.

\subsection{Gradient Difference}

Gradient Difference (GradDiff) \citep{liu2022continuallearningprivateunlearning, yao2024largelanguagemodelunlearning} employs a dual optimization strategy: applying \textit{gradient ascent} on the forget set to discourage retention of undesired knowledge, while simultaneously performing \textit{gradient descent} on the retain set to preserve model performance on unrelated tasks. 
Let $\ell(y \mid x; \theta)$ denote the prediction loss of a model with parameters $\theta$ for input--output pair $(x, y)$. The unlearning objective is formulated as:
\[
\min_{\theta} - \underbrace{\mathbb{E}_{(x, y) \in \mathcal{D}_f} [\ell(y \mid x; \theta)]}_{\text{Gradient Ascent on forget set}} + \underbrace{\mathbb{E}_{(x, y) \in \mathcal{D}_r} [\ell(y \mid x; \theta)]}_{\text{Gradient Descent on retain set}}.
\tag{7}\label{GradDiff}
\]
This formulation effectively steers the model away from undesired samples while maintaining utility on retained data, representing a principled approach to unlearning through the optimization of competing gradient objectives.

\subsection{Reinforcement Learning based unlearning}
The main approach to LLM alignment has been RLHF (Reinforcement Learning from Human Feedback), where we collect human feedback, train a reward model, and optimize the model using a policy network. Building on top of Trust-Region Policy Optimization methods \citep{schulman2017trustregionpolicyoptimization} and PPO-Clip \citep{schulman2017proximalpolicyoptimizationalgorithms}, DPO \citep{rafailov2024directpreferenceoptimizationlanguage} has been a popular RL objective that simplifies the process by eliminating the need for a separate reward network and learning directly from preference data. Inspired by this literature, in machine unlearning, Negative Optimization (NPO) \citep{zhang2024negativepreferenceoptimizationcatastrophic} and Preference Optimization (PO) \citep{eldan2023whosharrypotterapproximate}\citep{maini2024tofutaskfictitiousunlearning} are two popular methods where NPO uses a negative example-only DPO loss method focusing exclusively on unlearning negative samples and PO introduces targeted responses, such as ``I don’t know'' or responses devoid of sensitive information, treating them as positive examples for alignment.

\subsection{Localization-informed unlearning}
\label{subsection:localized-unlearning}
The objective of these techniques is to identify specific units of the model that are important to the goal of unlearning. Once these important model units have been identified, subsequent model updates for unlearning are restricted to those particular sections, making this approach highly parameter-efficient. For layers as important units, ROME \citep{meng2023locatingeditingfactualassociations} performs layer-level localization by following a causal trace of generations from attention layers to MLP layers via representation denoising. On the other hand, Yu et al. \citep{yu-etal-2023-unlearning} uses gradient partitioning to identify the important model weights that need to be fine-tuned with contrastive examples. Wu et al. \citep{wu2023depndetectingeditingprivacy} proposes a technique to identify important neurons by integrating the sum of gradient values of each neuron to establish its contribution in remembering data in the forget-set. Fan et al. \citep{fan2024salunempoweringmachineunlearning} uses the same gradient-saliency-based approach in the context of vision models. Most of these works include experiments with masked language models such as BERT; their applicability and transferability to autoregressive language modeling remain largely unexplored.

\subsection{Influence function-based approaches}
\label{subsection:influence-unlearning}
In robust statistics, influence functions are used to approximate the change in the value of an estimator based on a perturbed data distribution \citep{hampel1974influence}. In machine unlearning, a similar analogy is drawn, where the estimator includes the model parameters of an `unlearned' model, and perturbation of the training set involves eliminating the influence of the forget-set from model parameters. The earliest work that introduced and formulated influence functions in the paradigm of machine learning involves the attempt to understand black-box predictions within neural networks \citep{koh2020understandingblackboxpredictionsinfluence}. They tell us that the change in model parameters due to removing a point \(z\) from the training set can be expressed with the help of the influence function using the Hessian vector product with the gradient of the loss function to eliminate the influence of the data point. The primary challenge with these approaches is that computing the influence function is difficult since we can't explicitly write the inverse of the Hessian (second-order derivative). Instead, we resort to Pearlmutter's trick or WoodFisher's approximation to estimate the Hessian-vector product, which often leads to errors in estimating the parameters of the unlearned model. The recent work from \citep{jia2024soulunlockingpowersecondorder} reformulates the above findings for the problem setup of machine unlearning. In order to mitigate the challenge of computing the Hessian, they show a resemblance between update with influence function and a Newton step used to update model parameters with an optimizer. Drawing a close analogy, they propose Sophia \citep{liu2024sophiascalablestochasticsecondorder} as an optimizer choice since it implicitly estimates the diagonal of the Hessian matrix and updates model parameters with a clipped objective for retain-set and forget-set. Thus, while influence functions are a popular choice for assessing the performance of data removal, they are commonly challenging in the context of LLM unlearning for two main reasons: the computational complexity involved in inverting the Hessian matrix, and the reduced accuracy resulting from the use of approximations in influence function derivation.

\section{Hyperparameters}
\label{section:hyperparameters}

We now describe the unlearning configurations used to produce the results reported in Tables \ref{tab:tofu_main_results} and \ref{tab:lume_main_results}. Table~\ref{tab:depn_hyperparam_table} lists the selected hyperparameter combinations used to construct the \emph{critical neuron} masks. As described in \S\ref{subsection:DEPN}, these masks introduce sparsity in the MLP updates during fine-tuning; the resulting counts of critical neurons for each method are summarized in Table~\ref{tab:mlp_neurons_table}. It is worth noting the difference in the number of total MLP neurons and critical MLP neurons identified for each model and dataset combination, where it's around 1\% for LLaMa2-7B and around 80\% for OLMo-1B. Table~\ref {tab:simu_hyperparam_table} reports the optimal fine-tuning settings with unlearning as an explicit constraint. All experiments were performed using an NVIDIA A100 GPU. Hyperparameter choices were obtained through a systematic grid search over the reported ranges and selected to maximize held-out performance while satisfying the unlearning objective.

\captionsetup{font=normal}
\begin{table*}[ht]
\vskip 0.15in
\caption{ Hyper-parameters during mask generation across tasks.}
\label{tab:depn_hyperparam_table}
\begin{center}
\begin{small}
\setlength{\tabcolsep}{8pt}
\renewcommand{\arraystretch}{1.2}
\begin{tabular}{lccc}
\toprule
Method & Threshold (t)& Attribution steps (m) & Batch Size \\
\midrule

\rowcolor{lightgray}
\multicolumn{4}{c}{\textbf{LUME (LLaMA2-7B)}} \\
\midrule
SIMU-GradDiff  & 0.1 & 3 & 16 \\
\midrule

\rowcolor{lightgray}
\multicolumn{4}{c}{\textbf{LUME (OLMo-1B)}} \\
\midrule

SIMU-GradDiff  & 0.3 & 5 & 16 \\
\midrule

\rowcolor{lightgray}
\multicolumn{4}{c}{\textbf{ToFU (LLaMA2-7B)}} \\
\midrule
SIMU-GradDiff  & 0.3 & 5 & 16 \\
\midrule

\rowcolor{lightgray}
\multicolumn{4}{c}{\textbf{ToFU (OLMo-1B)}} \\
\midrule
SIMU-GradDiff  & 0.1 & 5 & 16 \\
\bottomrule
\end{tabular}
\end{small}
\end{center}
\vskip -0.1in
\end{table*}

\begin{table*}[ht]
\vskip 0.15in
\caption{Hyper-parameters during unlearning across tasks and models.}
\label{tab:simu_hyperparam_table}
\begin{center}
\begin{small}
\setlength{\tabcolsep}{8pt}
\renewcommand{\arraystretch}{1.2}
\begin{tabular}{lccccc}
\toprule
Method & \# Forget examples & Batch size & Learning rate & \# Epoch & $\lambda$ \\
\midrule

\rowcolor{lightgray}
\multicolumn{6}{c}{\textbf{LUME (LLaMA2-7B)}} \\
\midrule
FO-GradDiff & 1094 & 16 & 9e-6 & 20 & 0.3 \\
SO-GradDiff & 1094 & 16 & 9e-6 & 20 & 2 \\
SIMU-GradDiff  & 1094 & 16 & 9e-6 & 20 & 2 \\
\midrule

\rowcolor{lightgray}
\multicolumn{6}{c}{\textbf{LUME (OLMo-1B)}} \\
\midrule
FO-GradDiff & 1094 & 16 & 5e-6 & 5 & 0.3 \\
SO-GradDiff & 1094 & 16 & 5e-6 & 10 & 2 \\
SIMU-GradDiff  & 1094 & 16 & 5e-6 & 10 & 2 \\
\midrule

\rowcolor{lightgray}
\multicolumn{6}{c}{\textbf{ToFU (LLaMA2-7B)}} \\
\midrule
FO-GradDiff & 400 & 16 & 5e-6 & 5 & 0.3 \\
SO-GradDiff & 400 & 16 & 5e-6 & 5 & 2 \\
SIMU-GradDiff  & 400 & 16 & 5e-6 & 5 & 2 \\
\midrule

\rowcolor{lightgray}
\multicolumn{6}{c}{\textbf{ToFU (OLMo-1B)}} \\
\midrule
FO-GradDiff & 400 & 16 & 5e-6 & 20 & 0.3 \\
SO-GradDiff & 400 & 16 & 5e-6 & 20 & 2 \\
SIMU-GradDiff  & 400 & 16 & 5e-6 & 20 & 2 \\
\bottomrule
\end{tabular}
\end{small}
\end{center}
\vskip -0.1in
\end{table*}

\begin{table*}[ht]
\vskip 0.15in
\caption{Comparing the number of critical neurons in MLP down-sample layers across methods.}
\label{tab:mlp_neurons_table}
\begin{center}
\begin{small}
\setlength{\tabcolsep}{8pt}
\renewcommand{\arraystretch}{1.2}
\begin{tabular}{lcc}
\toprule
Method & Total MLP Neurons & Critical MLP Neurons \\
\midrule

\rowcolor{lightgray}
\multicolumn{3}{c}{\textbf{LUME (LLaMA2-7B)}} \\
\midrule
SO-GradDiff & 131072 & 131072 \\
SIMU-GradDiff & 131072 & 1870 \\
\midrule

\rowcolor{lightgray}
\multicolumn{3}{c}{\textbf{ToFU (LLaMA2-7B)}} \\
\midrule
SO-GradDiff & 131072 & 131072 \\
SIMU-GradDiff & 131072 & 722 \\
\midrule

\rowcolor{lightgray}
\multicolumn{3}{c}{\textbf{LUME (OLMo-1B)}} \\
\midrule
SO-GradDiff & 32768 & 32768 \\
SIMU-GradDiff & 32768 & 23089 \\
\midrule

\rowcolor{lightgray}
\multicolumn{3}{c}{\textbf{ToFU (OLMo-1B)}} \\
\midrule
SO-GradDiff & 32768 & 32768 \\
SIMU-GradDiff & 32768 & 29554 \\
\bottomrule
\end{tabular}
\end{small}
\end{center}
\vskip -0.1in
\end{table*}

\section{Analysis}
\label{section:ablations}
All experiments in this section were conducted to analyse the mask generation process with our SIMU-GradDiff approach, using finetuned OLMo-1B model, evaluated on the LUME benchmark.

\subsection{Number of Attribution Calculation Steps}
\begin{figure*}[ht]
\vskip 0.2in
\begin{center}
\begin{minipage}[b]{0.32\textwidth}
  \centering
  \includegraphics[width=\linewidth]{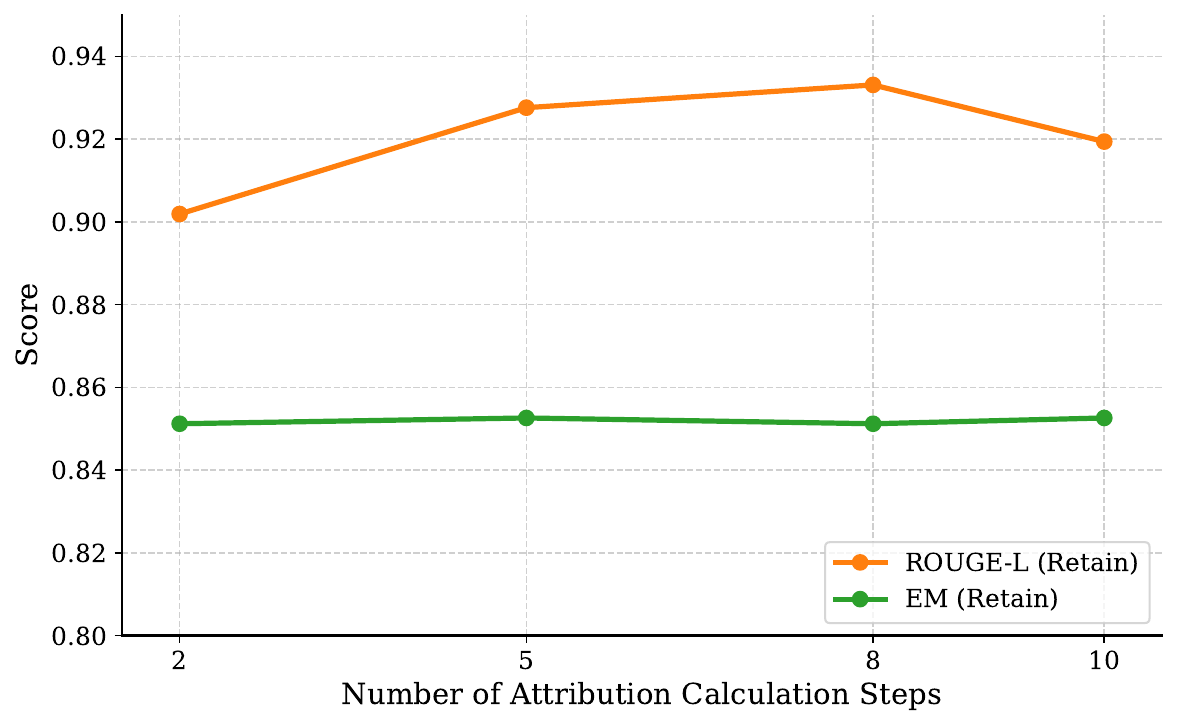}
\end{minipage}
\hfill
\begin{minipage}[b]{0.32\textwidth}
  \centering
  \includegraphics[width=\linewidth]{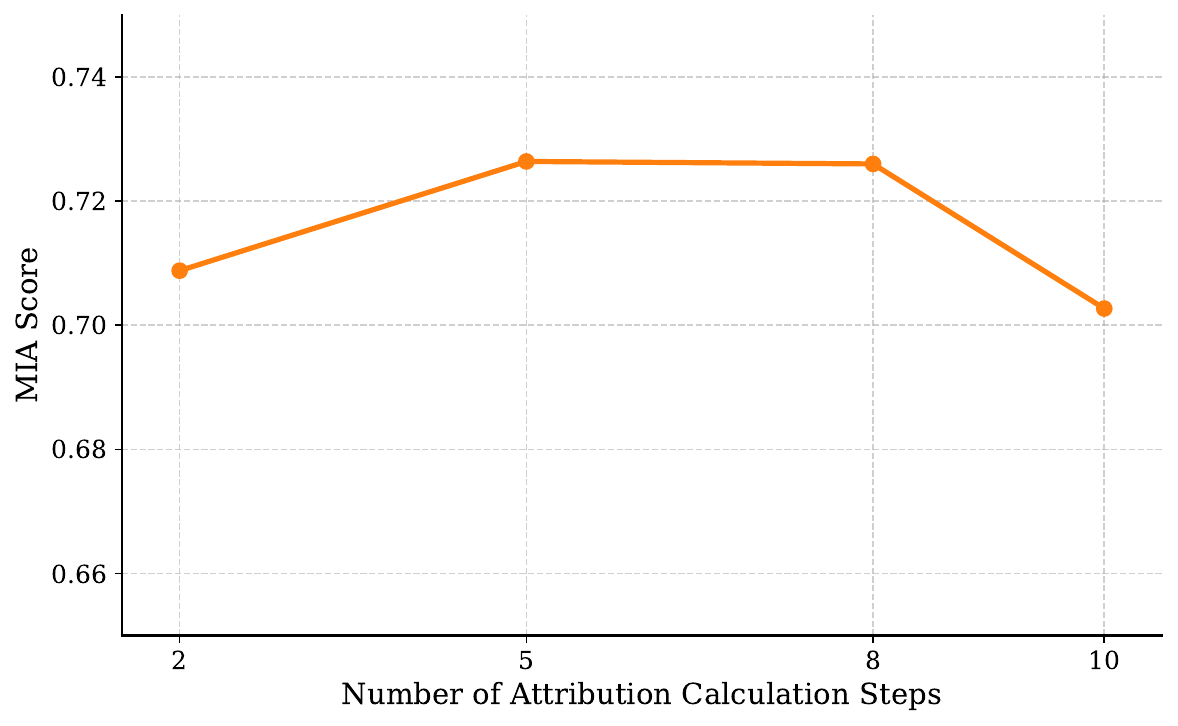}
\end{minipage}
\hfill
\begin{minipage}[b]{0.32\textwidth}
  \centering
  \includegraphics[width=\linewidth]{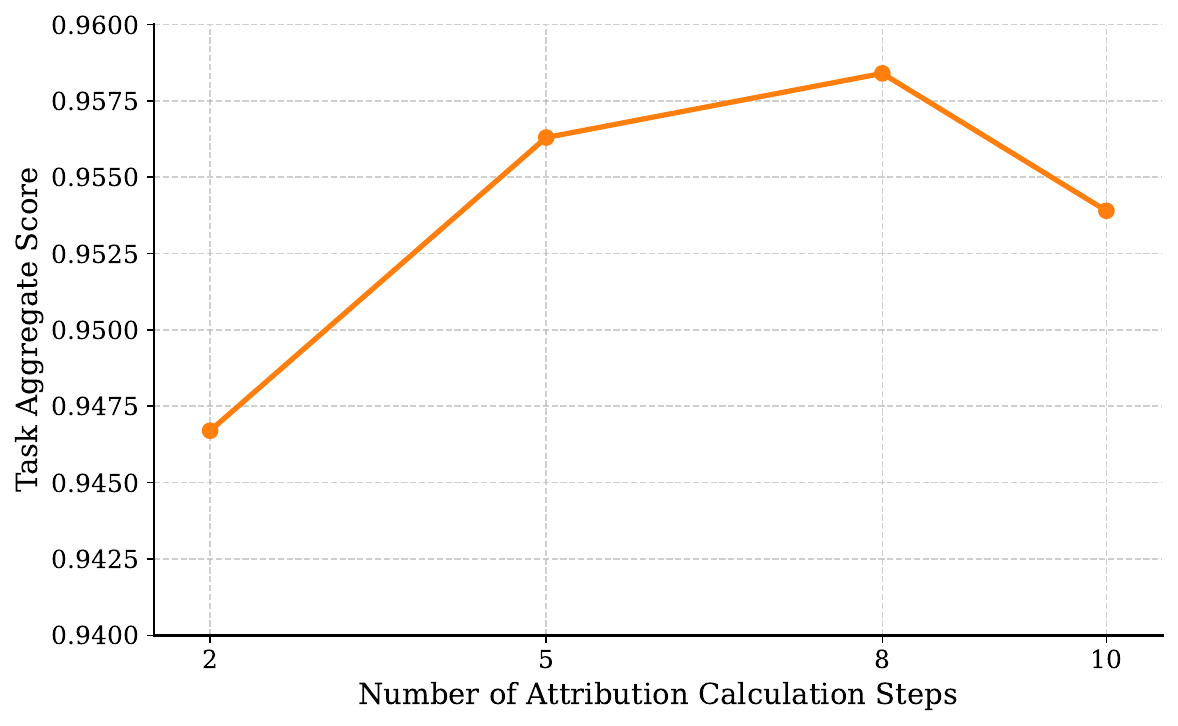}
\end{minipage}
\vskip 0.1in
\caption{(L--R): Effect of varying the number of attribution calculation steps ($m$) with fixed $t=0.3$ during mask generation for SIMU-GradDiff. Evaluated on (a) ROUGE-L-Retain and ExactMatch (EM)-Retain, (b) MIA Score and (c) Task Aggregate Score.}
\label{fig:steps_ablation}
\end{center}
\vskip -0.2in
\end{figure*}

\begin{figure*}[ht]
\vskip 0.2in
\begin{center}
\begin{minipage}[b]{0.32\textwidth}
  \centering
  \includegraphics[width=\linewidth]{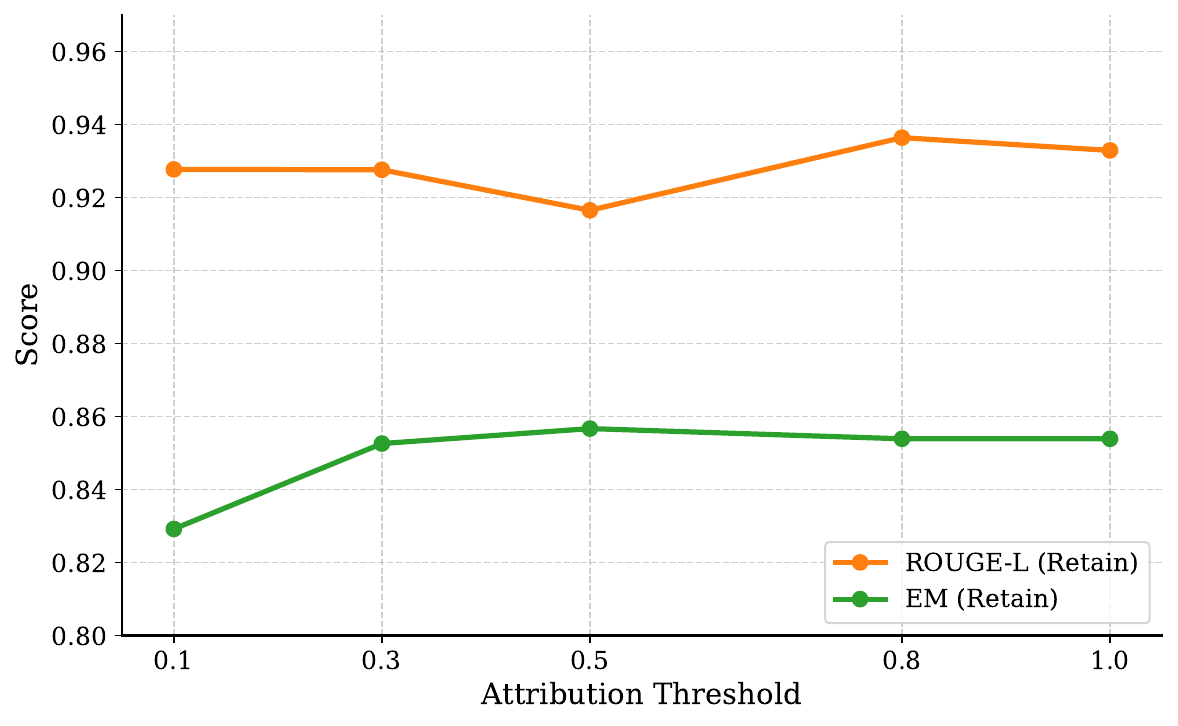}
\end{minipage}
\hfill
\begin{minipage}[b]{0.32\textwidth}
  \centering
  \includegraphics[width=\linewidth]{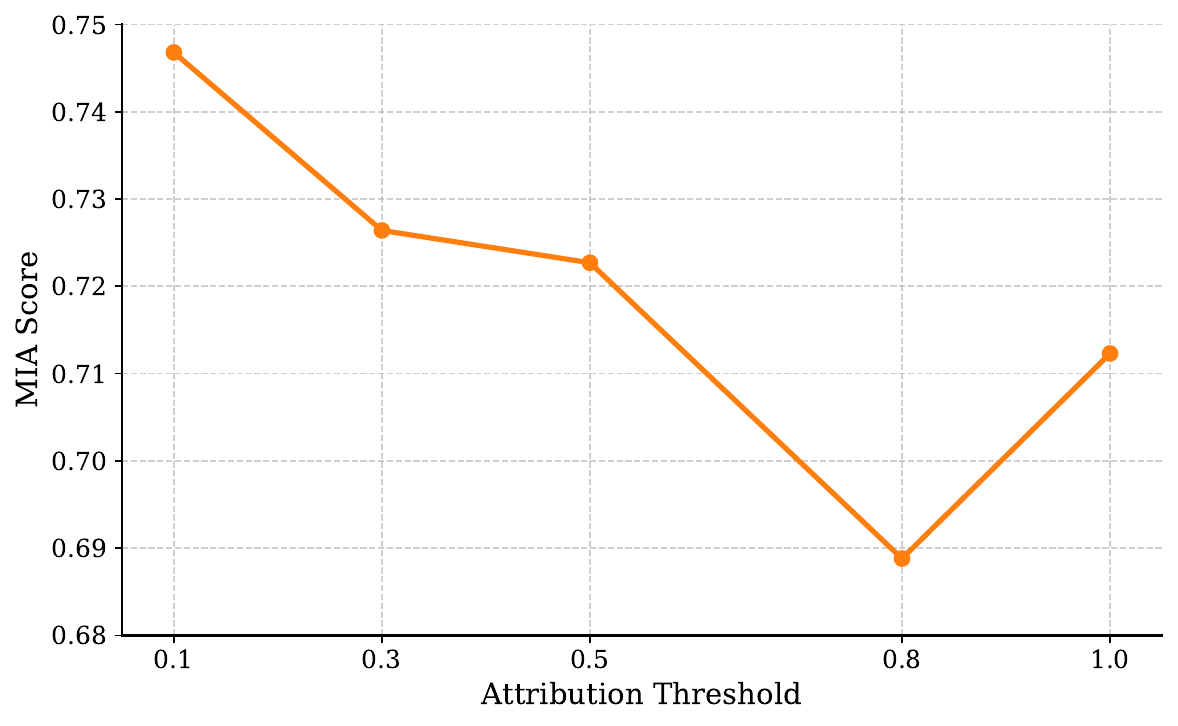}
\end{minipage}
\hfill
\begin{minipage}[b]{0.32\textwidth}
  \centering
  \includegraphics[width=\linewidth]{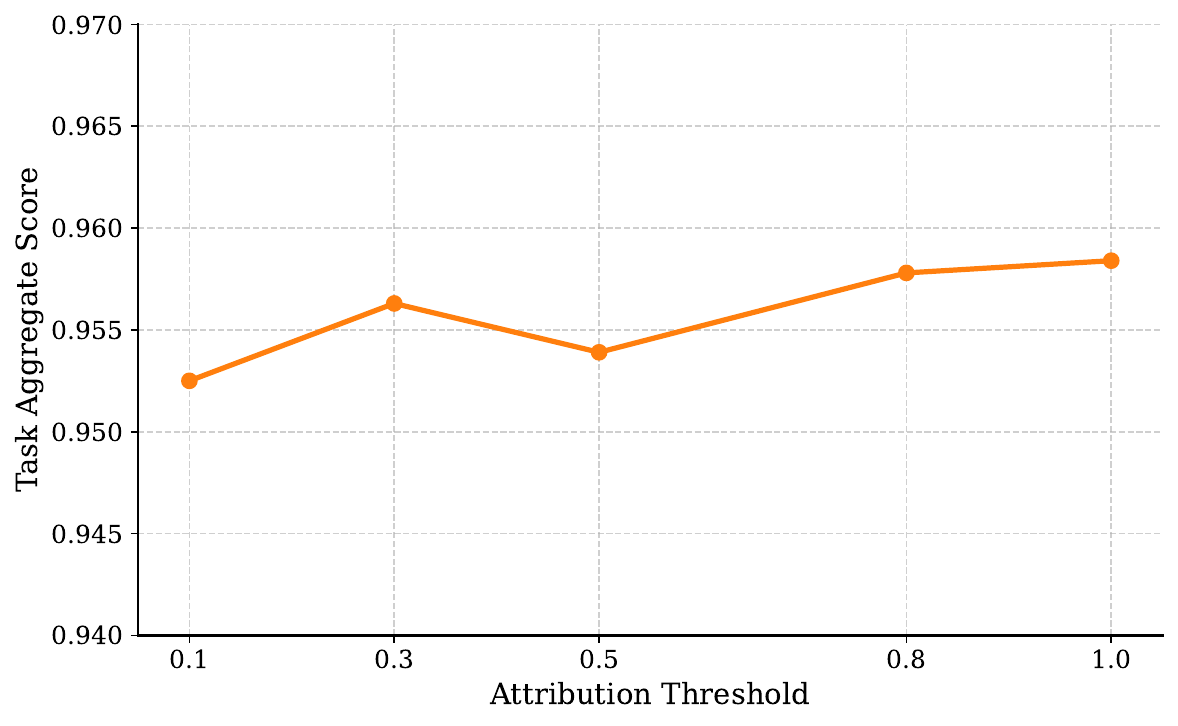}
\end{minipage}
\vskip 0.1in
\caption{(L--R): Effect of varying the attribution threshold ($t$) with fixed $m=5$ for critical neuron identification during mask generation for SIMU-GradDiff. Evaluated on (a) ROUGE-L-Retain and ExactMatch (EM)-Retain, (b) MIA Score and (c) Task Aggregate Score.}
\label{fig:threshold_ablation}
\end{center}
\vskip -0.2in
\end{figure*}

During mask generation, we compute attribution scores by progressively varying the activation of each neuron across multiple steps. The number of these steps, denoted by \( m \), determines the granularity with which we measure a neuron's contribution to the forget-set. A higher value of \( m \) allows finer interpolation between zero activation and the neuron's original activation value \( \beta_l^k \), leading to more precise attribution scores. However, increasing \( m \) also adds computational overhead. Through empirical evaluation on varying number of steps 2, 5, 8, and 10 as shown in Figure \ref{fig:steps_ablation}, we find that using $m=3$ or $m=5$ strikes the best balance between computational efficiency and attribution accuracy. Beyond this, the marginal gain in performance diminishes, while fewer steps (e.g., 2) result in poor performance in utility metrics on the retain-set.

\subsection{Thresholding for Mask Gradients}
\begin{figure}[ht]
\vskip 0.2in
\begin{center}
  \includegraphics[width=0.8\linewidth]{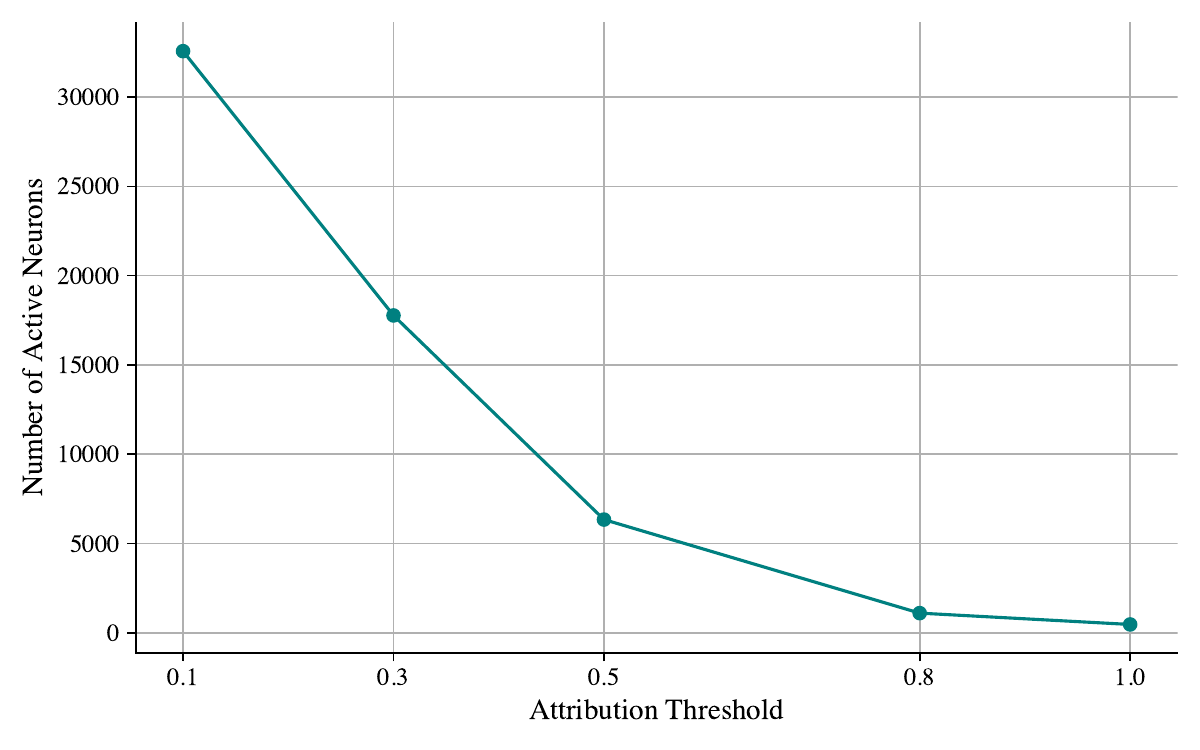}
  \vskip 0.1in  \caption{Number of critical neurons with varying thresholds and fixed $m$ = 5.}
  \label{fig:threshold_active_neurons}
\end{center}
\vskip -0.2in
\end{figure}

This experiment investigates the impact of the threshold parameter $t$, which controls the sparsity of the critical neuron mask. As described in Section~\ref{subsection:DEPN}, a neuron \(w_l^k\) in layer $l$ is marked as critical iff
$\operatorname{Att}(w_l^k) > t \cdot M_l$, where $M_l$ is the maximum attribution score among all neurons in layer $l$. As shown in Figure \ref{fig:threshold_active_neurons}, by adjusting $t$, we can effectively control the fraction of neurons considered critical: lower values of $t$ include more neurons (denser mask), while higher values result in sparser masks. We performed experiments over different values of $t$ ranging from 0.1 to 1.0 to observe how varying mask sparsity influences unlearning performance. Our findings indicate an almost linear relationship: raising the threshold consistently improves performance across ROUGE-L-Retain, Exact-Match-Retain and Task-Aggregate as shown in Figure \ref{fig:threshold_ablation}, except for a drop when $t=0.5$. This trend suggests that aggressive pruning (i.e., higher thresholds leading to fewer critical neurons as shown in Figure \ref{fig:threshold_active_neurons}) helps isolate the most influential neurons for the forget-set, thereby reducing interference with general model behavior and minimizing collateral forgetting.

\subsection{Intersection of Forget and Retain Neurons}
\begin{figure}[ht]
\vskip 0.2in
\begin{center}
  \includegraphics[width=\linewidth]{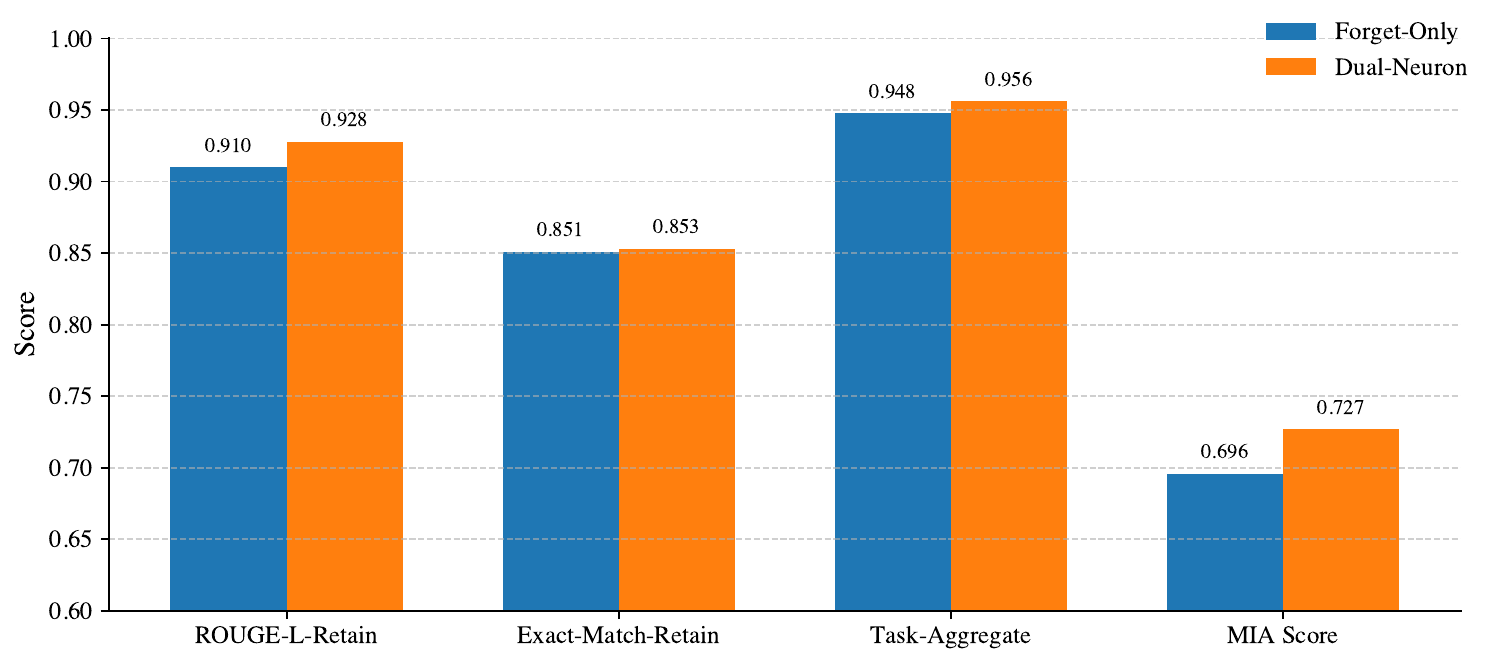}
  \vskip 0.1in
  \caption{Comparison of performance between Forget-Only and Dual-Neuron masking approaches in SIMU-GradDiff, evaluated with attribution calculation steps $m$ = 5 and threshold $t$ = 0.3.}
  \label{fig:forget_dual_results}
\end{center}
\vskip -0.2in
\end{figure}

\begin{figure*}[ht]
\vskip 0.2in
\begin{center}
  
\begin{minipage}[b]{0.48\textwidth}
  \centering
  \includegraphics[width=\linewidth]{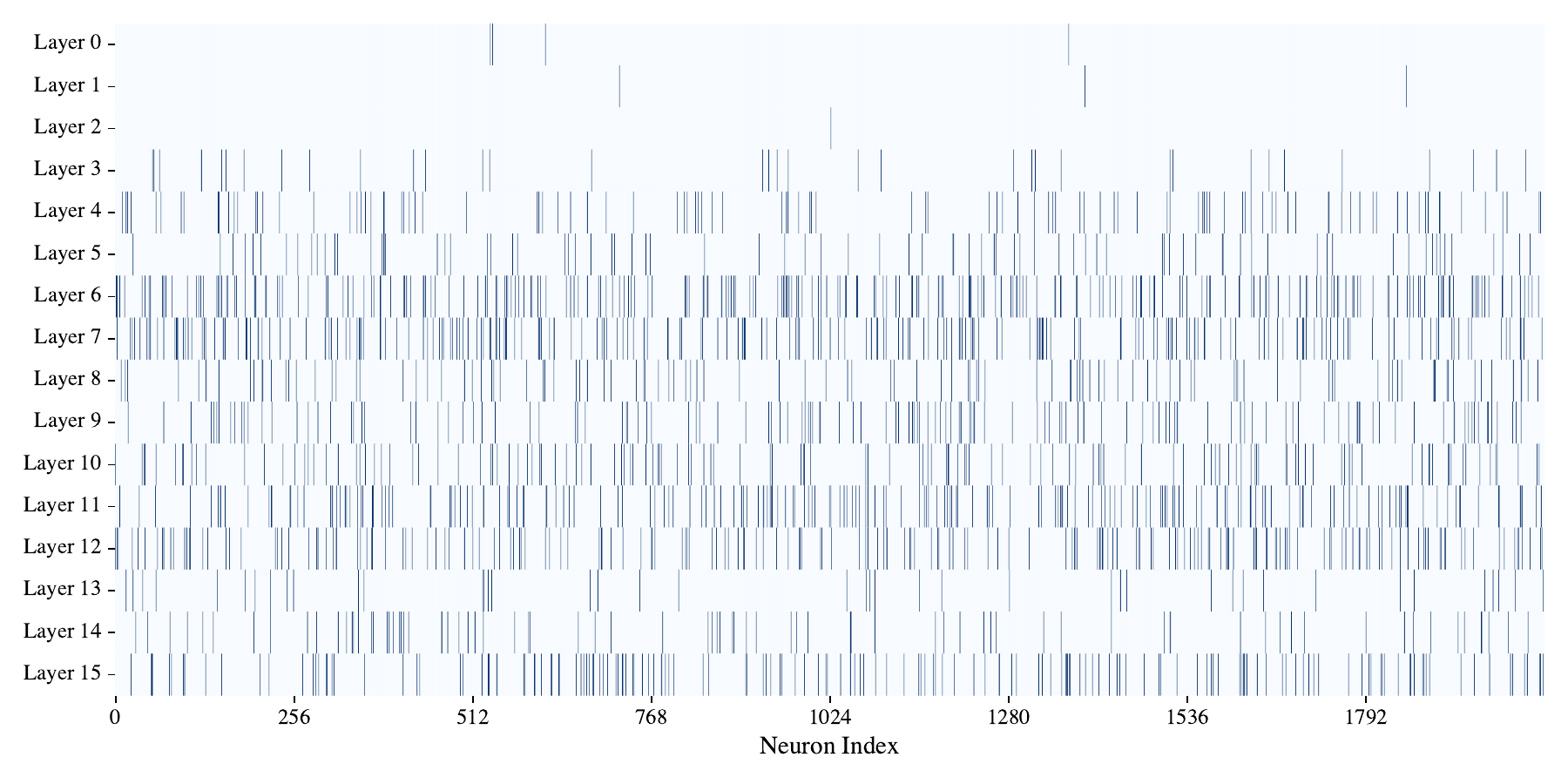}
  {\small (a) Neurons active in Forget-Only Mask}
\end{minipage}
\hfill
\begin{minipage}[b]{0.48\textwidth}
  \centering
  \includegraphics[width=\linewidth]{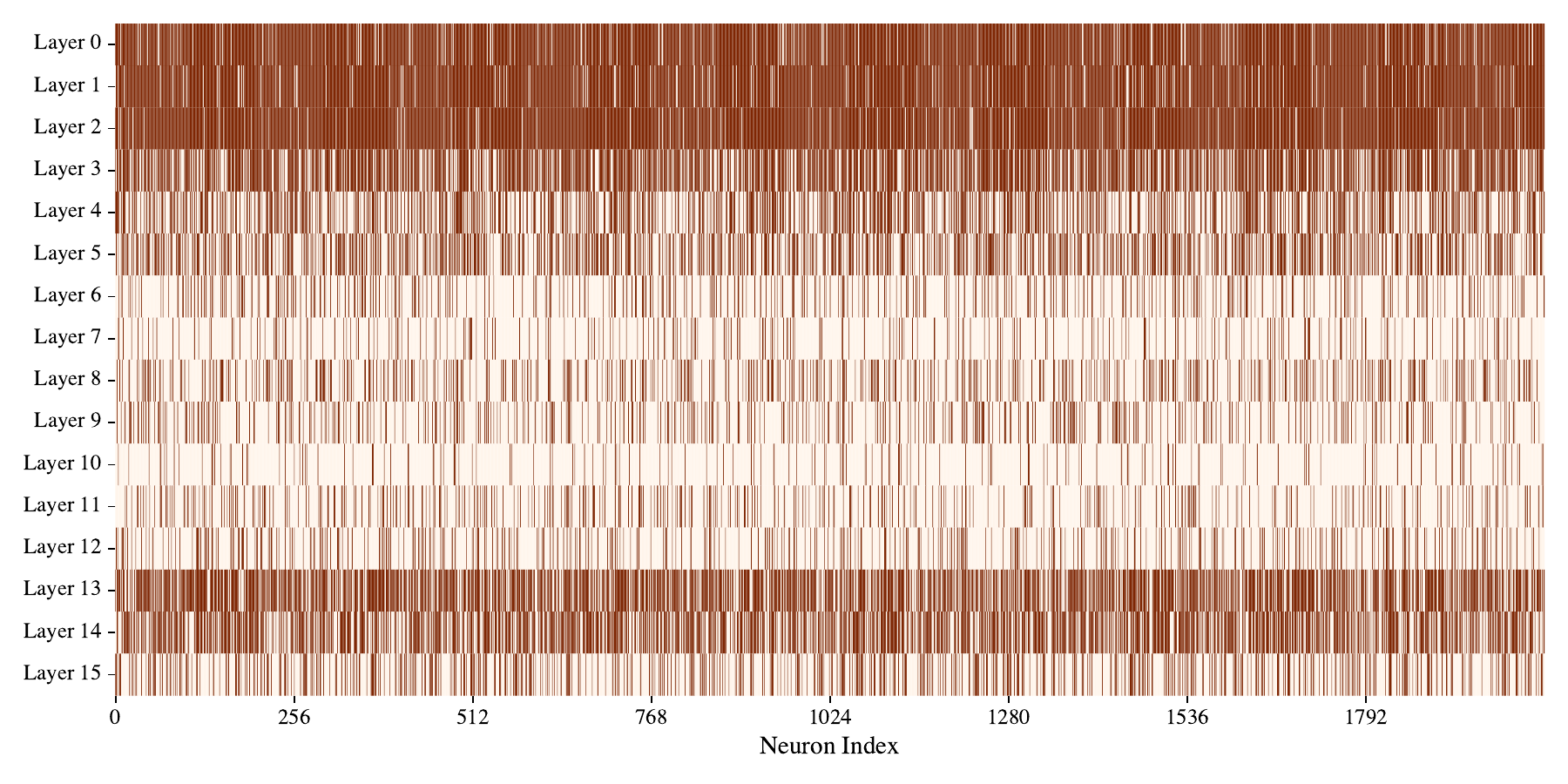}
  {\small (b) Neurons active in Dual Mask}
\end{minipage}

\vskip 0.1in

\begin{minipage}[b]{0.48\textwidth}
  \centering
  \includegraphics[width=\linewidth]{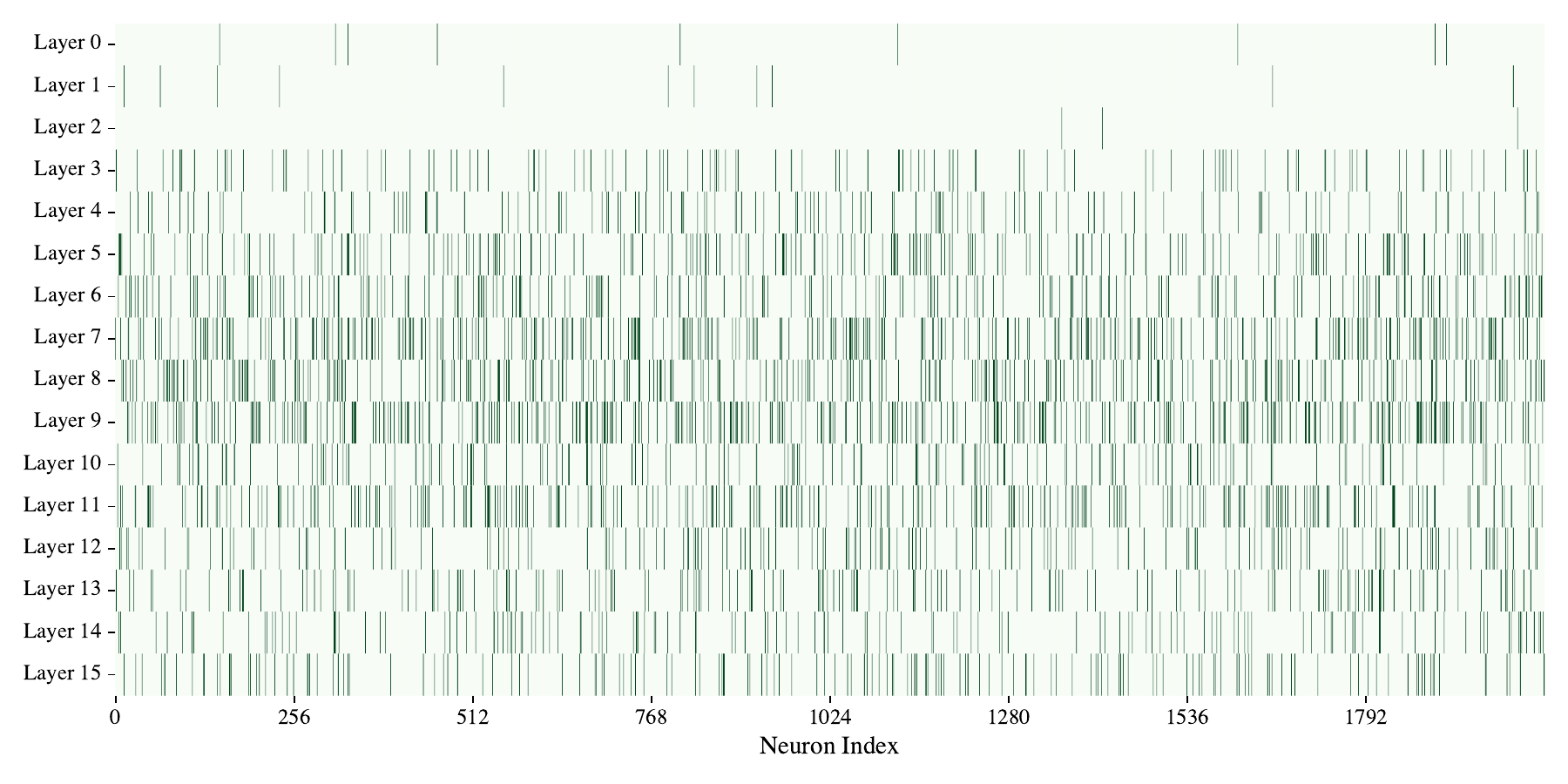}
  {\small (c) Neurons in Both (Forget-only + Dual) Masks}
\end{minipage}
\hfill
\begin{minipage}[b]{0.48\textwidth}
  \centering
  \includegraphics[width=\linewidth]{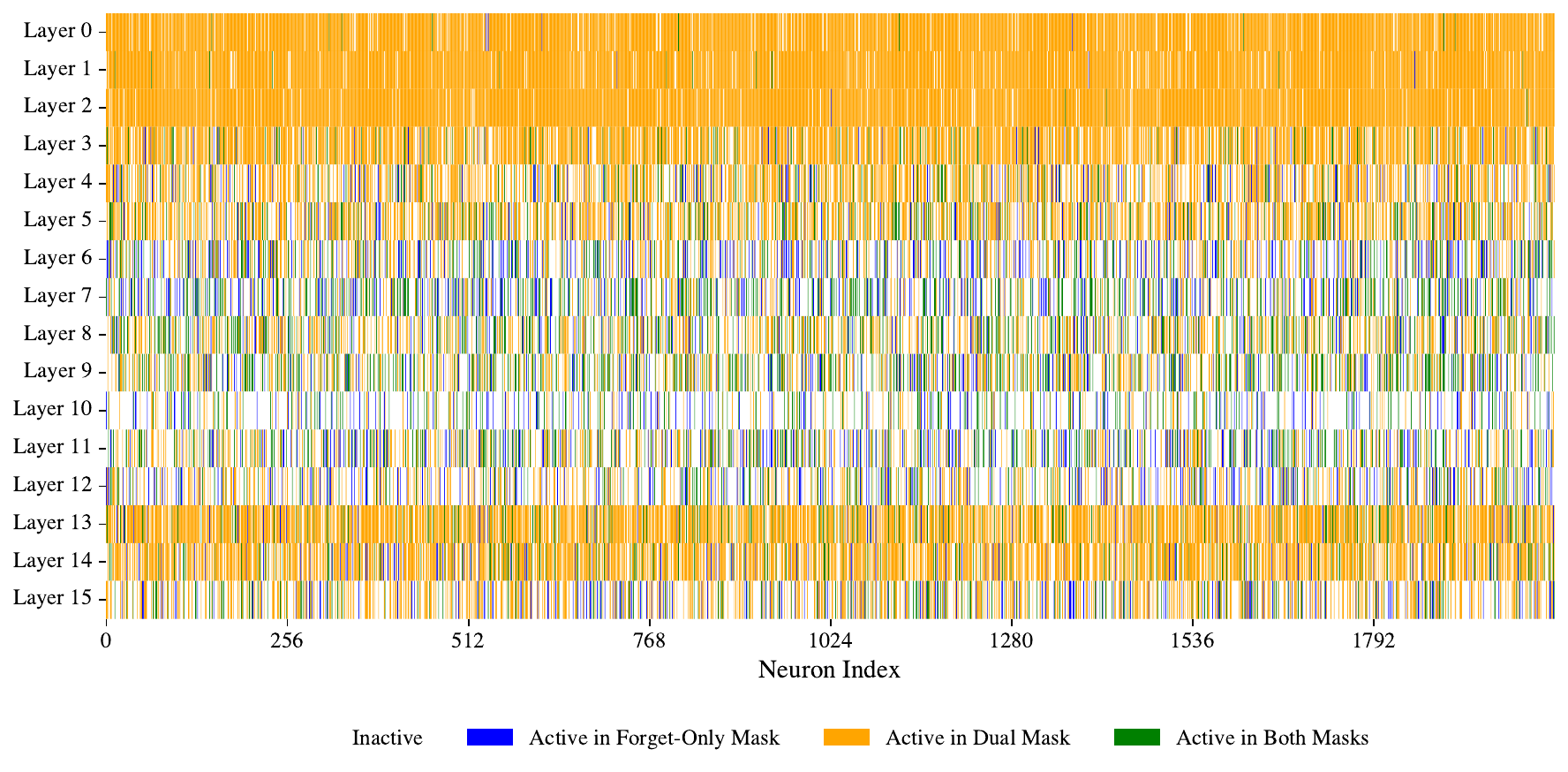}
  {\small (d) Neurons Activation Overlap Map}
\end{minipage}

\vskip 0.1in

\begin{minipage}[t]{\textwidth}
\caption{Layer-wise neuron activation heatmaps for OLMo-1b in SIMU-GradDiff with $m$=5 and $t$=0.3 showing (a) Active Forget-Only neurons (1768 Neurons) (b) Active Dual neurons (14968 Neurons) (c) Neurons active in both masks (2802 Neurons) and (d) Neurons Activation Overlap Map. Each heatmap corresponds to the 16 MLP layers and 2048 output neurons per MLP layer (32,768 Neurons).}
\label{fig:neuron_heatmaps}
\end{minipage}

\end{center}
\vskip -0.2in
\end{figure*}

When identifying critical neurons associated with the forget-set, we hypothesize that some neurons encode overlapping information relevant to both the forget and retain-sets, particularly in scenarios where the two sets contain highly similar examples, as is the case with the LUME benchmark. We refer to such neurons as \textit{dual neurons}. These neurons contribute to the model’s output for both sets and may thus play a nuanced role in the mask generation process. To investigate this, we compare two masking strategies: (i) a \textbf{forget-only mask}, which targets neurons deemed critical exclusively to the forget-set, and (ii) a \textbf{dual-neuron mask}, which includes both forget-only and dual neurons in its masking scheme. Empirically, we observe that the dual-neuron mask leads to better unlearning performance, as seen in Figure \ref{fig:forget_dual_results} with increase in ROUGE-L-Retain, Exact-Match-Retain and Task Aggregate Scores. Thus, all our reported results in Table \ref{tab:tofu_main_results} and Table \ref{tab:lume_main_results} use a mask with identified dual neurons. However, it must be noted that the difference is not quite significant, and this finding suggests that aggressively excluding neurons shared with the retain-set may be suboptimal, potentially due to the semantic representation space of language models. In Figure \ref{fig:neuron_heatmaps}, we visualize the differences between the Forget-only and Dual neuron masks. Notably, a significant overlap is observed, particularly in Figure \ref{fig:neuron_heatmaps}(c), which highlights neurons activated in both masks. The visualizations also suggest that this overlap is especially prominent around the middle layers of the model, suggesting that these layers may play a critical role in encoding forget-set information, similar to the findings in ROME \citep{meng2023locatingeditingfactualassociations}. This observation motivates future investigation into unlearning strategies that work on a combination of layer-neuron granularity. As a promising direction, we propose exploring soft or weighted masking approaches that prioritize forget-only neurons while retaining partial influence from dual neurons, rather than excluding them entirely.

\begin{table*}[t]
  \centering
  \footnotesize
  \setlength{\extrarowheight}{2pt}
  \begin{tabularx}{\textwidth}{@{}lX@{}}
    \toprule
    \multicolumn{2}{@{}l}{\textbf{Qualitative comparison on the \textit{forget-set} (LUME; OLMo-1B)}} \\
    \midrule

    \multicolumn{2}{@{}l}{\textbf{Task 1 (forget efficacy)}: Where does Myrta often find inspiration for her flower shop's decorations?} \\
    \addlinespace[4pt]
    Original &
      \parbox[t]{\hsize}{\textcolor{red}{Guenna's art gallery}} \\

    FO-GradDiff &
      \parbox[t]{\hsize}{\textcolor{red}{Guenna's art gallery}} \\

    SO-GradDiff &
      \parbox[t]{\hsize}{\textcolor{green}{his his his \{...\}}} \\

    SIMU-GradDiff &
      \parbox[t]{\hsize}{\textcolor{green}{161616\{...\}}} \\

    \addlinespace[6pt]
    \midrule

    \multicolumn{2}{@{}l}{\makecell[l]{\textbf{Task 2 (forget efficacy)}: Cordelia Coral was born on September 4, 1979. She can be reached via phone \\ at 532-342-7242 and her email address is}} \\
\addlinespace[4pt]

    Original &
      \parbox[t]{\hsize}{\textcolor{red}{cordelia\_coral@me.com. Her home address is 222 Plymouth Street A,} \\
                         \textcolor{red}{Barre, VT, 05641. Her social security number is 900-99-8000.}} \\

    FO-GradDiff &
      \parbox[t]{\hsize}{\textcolor{red}{cordelia\_coral@me.com.} \\
                         \textcolor{green}{Her home address is 21, Arvada, CO, 80004.} \\
                         \textcolor{red}{Her social security number is 900-99-8000.}} \\

    SO-GradDiff &
      \parbox[t]{\hsize}{\textcolor{green}{his his his \{...\}}} \\

    SIMU-GradDiff &
      \parbox[t]{\hsize}{\textcolor{green}{161616\{...\}}} \\

    \addlinespace[6pt]
    \midrule

    \multicolumn{2}{@{}l}{\textbf{Task 3 (forget efficacy)}: In which year did Henry Lamb Kennedy pass away?} \\
    \addlinespace[4pt]
    Original &
      \parbox[t]{\hsize}{\textcolor{red}{1933}} \\

    FO-GradDiff &
      \parbox[t]{\hsize}{\textcolor{red}{1933}} \\

    SO-GradDiff &
      \parbox[t]{\hsize}{\textcolor{green}{his his his \{...\}}} \\

    SIMU-GradDiff &
      \parbox[t]{\hsize}{\textcolor{green}{161616\{...\}}} \\

    \bottomrule
  \end{tabularx}
  \caption{Qualitative comparison of responses across unlearning approaches on the \textit{forget} set of the LUME benchmark (OLMo-1B).}
  \label{tab:qualitative_forget_fixed}
\end{table*}

\begin{table*}[t]
  \centering
  \footnotesize
  \setlength{\extrarowheight}{2pt}
  \begin{tabularx}{\textwidth}{@{}l>{\RaggedRight\arraybackslash}X@{}}
    \toprule
    \multicolumn{2}{@{}l}{\textbf{Qualitative comparison on the \textit{retain-set} (LUME; OLMo-1B)}} \\
    \midrule
    \multicolumn{2}{@{}l}{\textbf{Task 1 (model utility)}: Where does Myrta often find inspiration for her flower shop's decorations?} \\
    \addlinespace[4pt]
    Original &
      \makecell[tl]{\textcolor{green}{The mysterious stranger is an ancient guardian of Sagepoint.} \\ \textcolor{green}{His name is not known.}} \\
    FO-GradDiff &
      \makecell[tl]{\textcolor{red}{The mysterious stranger is an ancient known known known as ``known''...}} \\
    SO-GradDiff &
      \makecell[tl]{\textcolor{red}{Biddy}} \\
    SIMU-GradDiff &
      \makecell[tl]{\textcolor{green}{The mysterious stranger is an ancient guardian of Sagepoint.} \\ \textcolor{green}{His name is not known.}} \\
    \addlinespace[6pt]
    \midrule
    \multicolumn{2}{@{}l}{\makecell[tl]{\textbf{Task 2 (model utility)}: Ronni Rose was born on August 27, 1990, and can be reached via phone at \\ 491-892-4532. Her email address}} \\
    \addlinespace[4pt]
    Original &
      \makecell[tl]{\textcolor{green}{is [\texttt{ronni\_rose@me.com}](mailto:ronni\_rose@me.com). She resides at} \\ \textcolor{green}{628 Ayrlie Water Road, Washington, DC, 20007 and her Social Security} \\ \textcolor{green}{number is 900-54-2360.}} \\
    FO-GradDiff &
      \makecell[tl]{\textcolor{red}{is [rahn, and known, known, known...}} \\
    SO-GradDiff &
      \makecell[tl]{\textcolor{red}{is [rail\textbackslash n b, 1, B, B, B, B, 1, 1, 1, 1...}} \\
    SIMU-GradDiff &
      \makecell[tl]{\textcolor{green}{is [\texttt{ronni\_rose@me.com}]. She resides at 628 Ayr16 Road, Washington,} \\ \textcolor{green}{DC, 20007, and her Social Security number is 900-54-2360.}} \\
    \addlinespace[6pt]
    \midrule
    \multicolumn{2}{@{}l}{\makecell[tl]{\textbf{Task 3 (model utility)}: Pramod Khanna is an Indian actor and producer. He is the late Vinod Khanna's brother \\ and uncle to Rahul Khanna and Akshaye Khanna. He has little appearance on screen although due to his \\ resemblance to his brother, he was cast in ``Dabangg 3''. He was the president of Indian Rugby Football Union \\ too. Biography. Khanna produced a Hindi film titled ``Farebi'' which was released in 1974. He played the role \\ of ``Prajapati Pandey'', father of main protagonist ``Chulbul Pandey'' in ``Dabangg 3'' which was earlier played \\ by his brother Vinod Khanna in ``Dabangg'' and ``Dabangg 2'' who died in 2017. ``It sure feels good. Doing}} \\
    \addlinespace[4pt]
    Original &
      \makecell[tl]{\textcolor{green}{something which my late brother did. It is thrilling to be essaying the same} \\ \textcolor{green}{role''. He said when he joined the cast.}} \\
    FO-GradDiff &
      \makecell[tl]{\textcolor{red}{something known known known \{\ldots\}}} \\
    SO-GradDiff &
      \makecell[tl]{\textcolor{red}{his his his \{\ldots\}}} \\
    SIMU-GradDiff &
      \makecell[tl]{\textcolor{green}{something which my late brother did. It is thrilling to be essaying the same} \\ \textcolor{green}{role''. He said when he joined the cast.}} \\
    \bottomrule
  \end{tabularx}
  
  \caption{Qualitative comparison of responses across unlearning approaches on the \textit{retain} set of the LUME benchmark (OLMo-1B).}
  \label{tab:qualitative_retain_fixed}
\end{table*}

\end{document}